\documentclass{article}
\usepackage[nonatbib]{neurips_2020}

\usepackage{caption,subcaption}
\newcommand{\bl}[1]{{\color{black}{#1}}}

\usepackage{amssymb}

\usepackage{times}
\usepackage{epsfig}
\usepackage{graphicx}
\usepackage{amsmath}
\usepackage{amssymb}
\usepackage{xcolor}
\usepackage{bm}
\usepackage[linesnumbered,ruled,vlined]{algorithm2e}
\SetKwInput{KwInput}{Input}
\SetKwInput{KwOutput}{Output}
\usepackage{multirow}

\newcommand{\norm}[1]{\left\lVert#1\right\rVert}

\DeclareMathOperator{\Tr}{Tr}

\usepackage[pagebackref=true,breaklinks=true,colorlinks,bookmarks=false]{hyperref}
\usepackage[nameinlink]{cleveref}

\begin{document}

\title{Are socially-aware trajectory prediction models really socially-aware?}

\author{
    Saeed Saadatnejad$^{1,*}$ \And Mohammadhossein Bahari$^{1,*}$ \And Pedram Khorsandi$^{2,\dagger}$   \And Mohammad Saneian$^{2,\dagger}$  \And Seyed-Mohsen Moosavi-Dezfooli$^{3}$ \And Alexandre Alahi$^{1}$ \and\\ 
    $^1$EPFL \quad $^2${Sharif university of technology} \quad $^3$ ETH Zurich \\
    {\tt\small {\{saeed.saadatnejad, mohammadhossein.bahari\}}@epfl.ch}\ \ \\
}

\maketitle
\let\thefootnote\relax\footnotetext{\leftline{$^*$ Equal contribution as the first authors. }}
\let\thefootnote\relax\footnotetext{\leftline{$\dagger$ Equal contribution as the second authors. }}
\begin{abstract}
Our transportation field has recently witnessed an arms race of neural network-based trajectory predictors. While these predictors are at the core of many applications such as autonomous navigation or pedestrian flow simulations, their adversarial robustness has not been carefully studied. In this paper, we introduce a socially-attended attack to assess the social understanding of prediction models in terms of collision avoidance. An attack is a small yet carefully-crafted perturbations to fail predictors. 
Technically, we define collision as a failure mode of the output, and propose hard- and soft-attention mechanisms to guide our attack. Thanks to our attack, we shed light on the limitations of the current models in terms of their social understanding.
We demonstrate the strengths of our method on the recent trajectory prediction models. 
Finally, we show that our attack can be employed to increase the social understanding of state-of-the-art models.
The code will be made available. 
\end{abstract}

\section{Introduction}
Understanding the social behavior of humans is a core problem in many transportation applications ranging from  autonomous navigation (\textit{e.g.}, social/delivery robots \cite{bennewitz2002learning} or autonomous vehicles \cite{bouhsain2020pedestrian,bahari2021injecting}), to microscopic pedestrian flow simulations \cite{SEER2014212,robin2009specification}. For a robot to navigate among crowds safely or for an autonomous vehicle to drive in urban areas harmlessly, human behavior anticipation is essential. In particular, dealing with humans makes the problem safety-critical. For instance, a self-driving car's wrong prediction in a crosswalk can put a pedestrian's life in danger. Being a safety-critical problem raises the need for careful assessments of the trajectory prediction methods to mitigate the risks associated with humans. 
Consequently, the robustness properties of those methods, as one of the important assessment aspects, should be carefully studied.
\begin{figure}[t]
    \begin{center}
        \includegraphics[width=0.7\columnwidth]{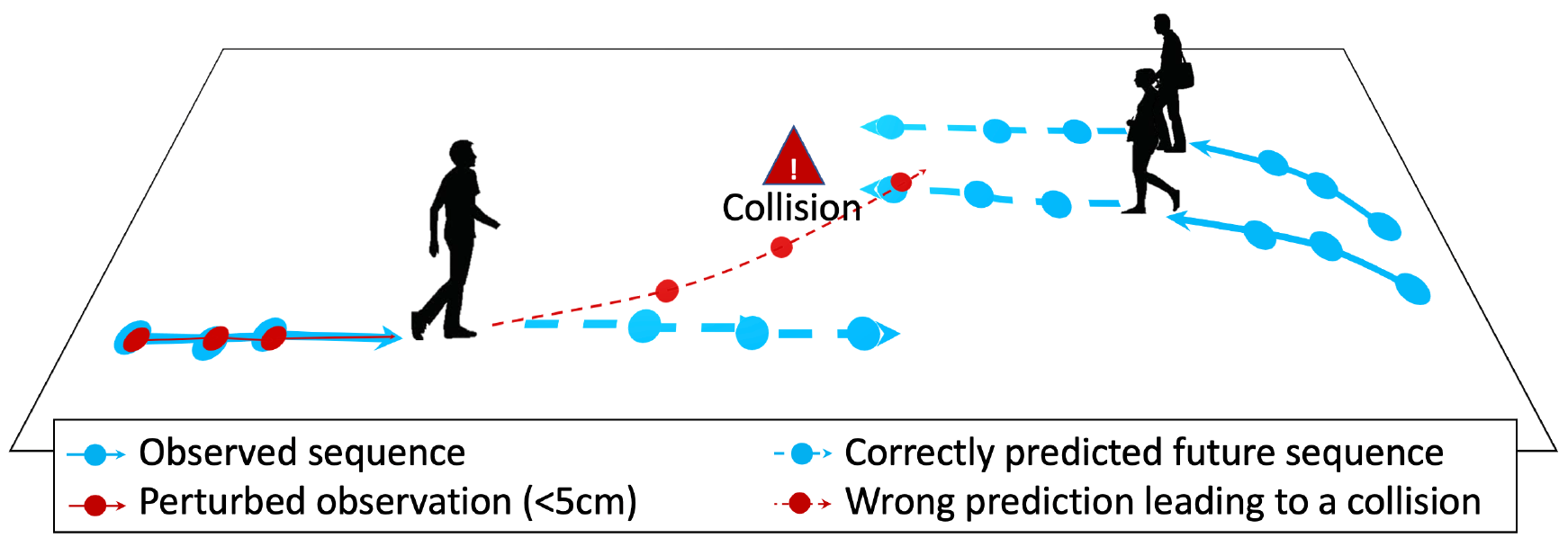}
    \end{center}
    \caption{Given the observation trajectories of the agents in the scene, a predictor (here S-LSTM \cite{alahi2016sociallstm}) forecasts the future positions reasonably (blue lines). However, with less than $5$~cm perturbation in the observation trajectory (in red), an unacceptable collision is predicted. }
\label{fig:pull}
\end{figure}

The pedestrian trajectory prediction problem is to predict future positions of pedestrians given their past positions as inputs.
The problem has received solutions based on microscopic human simulation models \cite{blue2001cellular,duives2013state}. Recently, diverse approaches based on neural networks are proposed. Various models based on 
Long-Short-Term-Memory networks \cite{alahi2016sociallstm}, convolutional neural networks \cite{nikhil2018convolutional}, and Generative Adversarial Networks (GAN) \cite{gupta2018socialgan} are presented. 
The main challenge of the problem  lies in learning the interactions between people.  Therefore, socially-aware neural network-based models are designed to tackle the interaction more accurately \cite{alahi2016sociallstm,vemula2018socialattention,kosaraju2019socialbigat,mohamed2020stgcnn}. 
Human interactions involve different social behaviors such as collision avoidance, walking within a group, and merging from different directions into a specific point. Among all behaviors, collision avoidance, \textit{i.e.,} people choosing a path that avoids collision with others, is one of the key behaviors rarely violated. That is why many previous works consider respecting collision avoidance as the evidence of their model being social \cite{mohamed2020stgcnn,kosaraju2019socialbigat,kothari2020human}. Similarly, we consider collision avoidance as an indicator of social behavior of the models.

We show a conceptually plausible real-world scenario in \Cref{fig:pull}. Given the observed trajectories of humans in the scene, a social predictor forecasts the future positions reasonably without collision.
However, by adding a small perturbation of less than $5$~cm to the observation trajectory, unexpectedly, a collision between predictions of agents occurs which indicates a non-complete social understanding by the predictors.
The trajectories in that figure comes from S-LSTM \cite{alahi2016sociallstm}.

In this work, in contrast to the common adversarial attacks which are designed for classifiers \cite{NEURIPS2020_1ea97de8,szegedy2013intriguing}, we design attack for the trajectory prediction problem which is a multimodal regression task.
We use adversarial examples to study the collision avoidance behavior of the trajectory prediction models.
More specifically, we investigate the worst-case social behavior of a prediction model under small perturbations of the inputs. 
This study has two primary motivations; 
(1) it is an evaluation method for the previously-proposed predictors. Our method brings counter-examples in which the models fail in having social behavior, \textit{i.e.,} it cannot avoid collision. (2) leveraging adversarial examples, one can train models with better collision-avoidance. Furthermore, our study highlights practical concerns for employing such models in real-world applications. Notably, it is shown that state-of-the-art localization algorithms give on-average more than $0.2$~m errors on human location detection at each frame \cite{deng2020joint,bertoni2020monstereo}. 
While our work focuses on model failures under adversarial settings, it motivates further studies of the model's performance when localization algorithms' error distributions are concerned.

We propose an adversarial attack to fool/fail the trajectory prediction models by causing collision between two agents' \textit{predicted trajectories}.
Namely, the adversarial attack aims at finding small perturbations that lead to a collision.
The collision can hypothetically happen between any two agents and at any prediction timestep. However, from the attack algorithm's perspective, the choice of the agent and the timestep impacts the final perturbation's size significantly. \color{black}
To address that, we introduce an attention-guided adversarial attack, named \textit{Socially-ATTended ATTack (S-ATTack)}, which learns the best collision points.
We present two variants of our attack: hard-attention and soft-attention. Our experiments demonstrate that our novel attack can find perturbations that make state-of-the-art trajectory prediction models generate wrong predictions, leading to collisions with small perturbations. We show that the achieved perturbations for one predictor can be transferred to other predictors revealing common weaknesses among models. Also, we demonstrate that the models are over-dependant on the last observation points which makes the models vulnerable. Lastly, we introduce an adversarial training scheme to make trajectory prediction models more robust. In particular, we show how our method can improve the models' social understanding in terms of collision avoidance. To the best of our knowledge, this is the first work addressing the adversarial vulnerability and robustness of trajectory prediction models.
Our main contributions are summarized as follows:
\begin{itemize}
    \item We introduce S-ATTack to assess the social understanding of the state-of-the-art trajectory prediction models.
    \item 
    Our experiments shed light on the weaknesses of prediction models from different aspects.
    \item We demonstrate how to improve the robustness properties of the predictors using our S-ATTack. 
\end{itemize}

\section{Related work}

\noindent\textbf{Human trajectory prediction.} 
Social-force model \cite{helbing1998social-forces} was one of the key hand-crafted methods proposed to capture human social behaviors. They model interactions between pedestrians by means of social fields determined by repulsive and attractive forces. 
\bl{Social interactions have been addressed from other perspectives} such as discrete choice modeling \cite{antonini2006discrete}, and cellular automaton model \cite{burstedde2001simulation, vizzari2015agent}. While all these hand-crafted methods have nice interpretability and are not data-heavy, it is shown that they are not able to effectively model long-term dependencies or complex interactions \cite{alahi2017sociallstmbook}. 

Social-LSTM \cite{alahi2016sociallstm} was the first work that proposed the use of data-driven neural networks instead of hand-crafted functions to learn interaction dynamics between agents in the scene. Many works pursued the use of deep learning and proposed diverse approaches for learning interactions. Different types of pooling social features are studied in \cite{kothari2020human,bartoli2018context,ivanovic2019trajectron,alahi2016sociallstm} to share features of agents leading to a social-compliant prediction.
Convolutional neural networks were studied to process pooled features of agents in the scene \cite{nikhil2018convolutional}.
In order to better detect relative importance of each agent, attention networks were employed in \cite{vemula2018socialattention}. 
Recently many works leveraged graph neural networks to model relations between agents \cite{kosaraju2019socialbigat,li2020social,yu2020spatio,mohamed2020stgcnn}. The multimodal distribution of data is learned by using generative adversarial networks  \cite{gupta2018socialgan,sadeghian2019sophie,amirian2019social}.
While all mentioned works improved the performance on average and final displacement errors, they occasionally output unacceptable solutions (e.g., collisions). In this work, we attack the state-of-the-art deep learning-based recent models and reveal their weaknesses against small perturbations which challenges their social behavior.

Previously mentioned works try to learn social behavior by observing human data, which implicitly include collision avoidance. There also exist works that explicitly teach models to avoid collisions. While most of these works address the planning problem \cite{zeng2019end,chen2017decentralized}, some adopted the inverse reinforcement learning framework to guide the network towards collision-free trajectories \cite{HeidenBMVC2019}. 
In this work, we focus our attack on the deep learning-based trajectory predictors which are fully data-driven and do not leverage human guidance. 

\noindent\textbf{Adversarial attacks.} 
Adversarial attack was first introduced in \cite{szegedy2013intriguing} by showing how vulnerable deep image classifiers are against non-perceptible yet carefully-crafted perturbations. Formally, an adversarial perturbation is defined as the minimal perturbation $R$ which changes the output of the given classifier $f$:
\begin{equation}
    \min_R ||R||_2 \: \: \text{subject to} \: \:  f(X+R) \neq f(X),
\end{equation}
where X is the input image. 
It is shown that adding these imperceptible perturbations bounded in terms of $\ell_p$ norms can easily fool image classifiers \cite{goodfellow2014explaining, moosavi2016deepfool, su2019onepixelattack}. 
According to the "human vision knowledge", adding imperceptible perturbations to an image does not change its category label. Therefore, a good model should preserve its prediction while adding these perturbations to its input.

Adversarial attacks are also generalized to assess vulnerability of models against perturbations in other domains such as natural language processing \cite{zhang2020adversarial,behjati2019universal} and tabular data  \cite{ballet2019imperceptible}. Nevertheless, to the best of our knowledge, adversarial attacks are not yet explored in the context of human trajectory prediction. Human trajectory is a temporal trajectory, hence can be seen as time-series data prediction. Different types of time-series data are studied in \cite{fawaz2019adversarial,karim2020adversarial,liu2020adversarial} in classification problems. 
While their focus is on the classification task, we target a regression task which makes the definition of a wrong output challenging.
In addition, in human trajectory sequences, the imperceptibility of the perturbations is not the main interest because here, human knowledge is to respect social behaviors (avoiding collisions) in all scenarios.
Lastly, a specific challenge to our problem is the freedom in the choice of the collision point which affects the perturbation size. We address these challenges by the proposed S-ATTack approach. 

\section{Method}
\label{sec:method}
In this section, we first explain the notations and definitions. Then, we will provide the details of S-ATTack. 

\subsection{Formulation}
\label{sec:background}
\subsubsection{Human trajectory prediction}
\label{sec:traj_pred}
Pedestrian trajectory prediction addresses a regression task with sequences as inputs and outputs.
At any timestep $t$, the $i$-th person/agent is represented by his/her xy-coordinates $(x_t^i, y_t^i)$.
We denote each agents' observation sequence for $T_{obs}$ timesteps as $X^i$, a $T_{obs} \times 2$ matrix and the observation sequences of all the $n$ agents in the scene as $X=(X^1, \dots, X^{n})$.
Given $X$, the trajectory predictor $f$ predicts the next $T_{pred}$ positions of all agents $Y=(Y^1, \dots, Y^{n})= f(X^1, \dots, X^{n})$ where $Y^i$ is a $T_{obs} \times 2$ matrix. 

\subsubsection{Adversarial examples for trajectory prediction}

Equipped with the notations introduced in \cref{sec:traj_pred}, we will provide a definition of adversarial examples for trajectory prediction.
In this paper, without loss of generality,
we assume the perturbation $R$ is added to the candidate agent which is arbitrarily chosen among agents. Note that in the experiments, all agents are considered as the candidate agent one by one.
$\hat{X}^1 = {X}^1 + R$ while the observations of other agents (which we refer to as neighbors) $\{X^j\}_{j \neq 1}$ remain fixed. Therefore, $R$ is a $T_{obs} \times 2$ matrix of adversarial perturbation, the adversarial example is $\hat{X}=(X^1 + R, X^2, \dots, X^{n})$ and the output of the predictor for that example is $\hat{Y}=(\hat{Y}^1, \dots, \hat{Y}^{n})=f(\hat{X})$.
Formally, given a small constant $\epsilon > 0$, a collision distance threshold $\gamma$ and the maximum of the norm of all rows of a matrix $\norm{\cdot}_{\max}$ , a socially-attended adversarial example is obtained if:
\begin{equation}
\begin{split}
\bl{\exists  t, j \neq 1, \norm{R}_{\max}\leq\epsilon
: \ \norm{\hat{Y}_t^j - \hat{Y}_t^1} < \gamma.} 
\end{split}
\label{eq:formulation2}
\end{equation}

In other words, this type of adversarial examples is based on perturbing an observation trajectory so that $f$ predicts the future timesteps with at least a collision (the distance less than $\gamma$) between two agents $j$ and $1$ in one timestep $t$. 
In addition, without loss of generality, we focus on the collisions between the candidate agent and neighbors. Clearly, it can directly be expanded to collisions between any two agents.
In the next section, we will describe how we obtain $R$ using Socially-attended attack.

\subsection{Socially-attended attack (S-ATTack)}
\label{sec:socially_attended}

We propose three optimization problems based on different attention mechanisms for socially-attended attack to find suitable perturbations for a collision. The optimization problems are explained in sections \ref{sec:naive}, \ref{sec:hard} and \ref{sec:soft} and are solved using an iterative algorithm based on projected gradient descent elaborated in \cref{sec:pgd}.

Before introducing the three optimizations, we will explain the distance matrix which is used in all three optimizations. 
Given the perturbation $R$, and a model $f$, we define the distance matrix $D(R) \in\mathbb{R}^{{(n-1)} \times T_{pred}}$ as a function of the input perturbation $R$. It includes the pairwise distance of all neighboring agents from the candidate agent in all prediction timesteps.
Let $d_{j,t}$ denote the element at $j$-th row and $t$-th column of $D(R)$, i.e., the distance of the agent $j$ from the candidate agent at timestep $t$ of the prediction timesteps:
\begin{equation}
   d_{j,t} := \norm{\hat{Y}_t^j - \hat{Y}_t^1}.
    \label{eq:D}
\end{equation}
Hence, for a particular $R$, the distance matrix $D(R)$ can be leveraged to indicate whether a collision has occurred. We now explain three methods to find such a perturbation by optimizing different cost functions depending on $D(R)$.

\subsubsection{No-attention loss}
\label{sec:naive}

Our first attempt is to introduce a simple social loss to find perturbations that make collisions in the trajectory prediction sequences.
To achieve that, we find the perturbations that make a collision among the predictions of humans in the prediction model.
One naive solution is to minimize the sum of distances between the candidate agent and its neighbors in all prediction timesteps: 
\begin{equation}
    \begin{split}   
        \min_R & \norm{D(R)}_F.
    \end{split}
    \label{eq:naive}
\end{equation}
 
This naive scheme gives the same attention to all agents, which may not be efficient in obtaining small perturbations. For instance, a far agent may not be a good potential candidate for collision and should receive lower attention.

\subsubsection{Hard-attention loss}
\label{sec:hard}
A better approach to cause a collision is to target a specific agent in a specific timestep instead of taking an average over all neighboring agents.
Then, the model is attacked in a way that the distance of the chosen agent with the candidate agent is decreased in the corresponding timestep until the collision occurs.
The equation of the hard-attention attack is as follows:
\begin{equation}
\begin{split}
    \min_{R,W}  &\Tr \left( W^\top D(R) \right) +  \lambda_r \norm{R}_F, \\
    \text{s.t. }&w_{j,t}=\delta_{jk}\delta_{tm},\quad
    k, m = \underset{j,t}{\text{argmin }}d_{j,t},
\end{split}
\label{eq:hard}
\end{equation}
where $\delta$ is the Kronecker delta function.
Besides, $W$ is the attention weight matrix and $w_{j,t}$ is the attention weight for the agent $j$ at timestep $t$ of the prediction timesteps. 
\bl{Indeed, the associated weight to the target agent $w_{k,m}$ which is the closest agent-timestep is $1$ and others are $0$.}
The socially-attended loss is the trace ($\Tr$) of multiplication of $D(R)$ by the transpose of $W$ added with
the regularization on the perturbation with the balancing coefficient $\lambda_r$ that encourages finding a small perturbation sequence to make a collision. 

\subsubsection{Soft-attention loss}
\label{sec:soft}
The main drawback of~\cref{eq:hard} is that the target point $(k, m)$ for a collision is selected based on the assumption that attacking the closest agent-timestep requires small-enough perturbation.
\bl{This confines the target point selection and might not find the most optimum target point. Note than the models are non-linear and a collision with the closest agent-timestep may not essentially require the smallest perturbation.}

To address that, we let the attack attend to the optimal target by itself.
We introduce a soft-attention mechanism in which the weights associated to each agent-timestep is assigned by the attack in order to achieve a smaller perturbation.
The equation of the soft-attention attack is as follows:
\begin{equation}
\begin{split}
    \min_{R,W}  &\Tr \left( W^\top \tanh(D(R)) \right) +  \lambda_r \norm{R}_F - \lambda_{w} \norm{W}_F, \\
    \text{s.t. }& \sum_{j,t}{w_{j,t}} = 1, \quad w_{j,t} \geq  0, \\
\end{split}
\label{eq:soft}
\end{equation}
where $\tanh$ is applied to the entries of $D(R)$ in order to concentrate less on very far agent-timesteps. Also, we discourage uniformity of weights by subtracting the Frobenius norm of W multiplied by a scalar $\lambda_w$.

$W$ is initialized with a uniform distribution. It is progressively updated and puts more weights to the more probable targets for making a collision. Near the convergence point, the best target agent receives a weight value close to $1$ while the rest receive $0$.
Note that $W$ and $R$ are optimized jointly per input sample.
We show one example of how $W$ changes in the training in the supplementary material.

We will compare our social adversarial attacks (hard-attention \cref{eq:hard} and soft-attention \cref{eq:soft}) in~\cref{sec:ablation}.
For brevity, in the rest of the paper, by S-ATTack, we refer to the attack with soft-attention.


\subsection{S-ATTack algorithm}
\label{sec:pgd}
In this section, we explain how adversarial perturbations are achieved using the introduced loss functions. The pseudo-code of the algorithm is written in \cref{alg:alg}. The method is an iterative algorithm with maximum $k_{max}$ iterations. At each iteration,
first, perturbed inputs $\hat{X}$ are calculated by adding perturbation $R$ to $X^1$. Then, we find new predictions $\hat{Y}$ using the perturbed inputs $\hat{X}$. Next, we use projected gradient descent \cite{madry2017pgd} to solve the constraint optimization problems introduced in the previous subsections. Namely, the perturbation $R$ is updated using the gradient of the equations defined above with hyperparameter $\alpha$ and then projected to the $\ell_{\infty}$ ball with radius $\epsilon$. Finally, if a collision in the predictions exists, the algorithm stops otherwise it continues until the maximum iterations. 

\begin{figure}[ht]
  \centering
  \begin{minipage}{0.59\linewidth}
\begin{algorithm}[H]
\DontPrintSemicolon
\SetAlgoLined
  \KwInput{Sequence $X$, Predictor $f$} 
  \KwOutput{Perturbed sequence $\hat{X}$}
  Initialize $k\gets 0$, $\hat{X}\gets X$\ ,$R\gets 0$ \;
\While{$k<k_{max}$}{
    $\hat{X}=(X^1 + R, X^2, \dots, X^{n})$ \;
    $\hat{Y} = f(\hat{X})$ \;
    $R = R + \alpha\nabla (\cref{eq:naive} \; or \;  \cref{eq:hard} \; or \;  \cref{eq:soft})$ \;
    $[R_{i,j}] = [$min$(R_{i,j}, \epsilon)$] \;
    compute $D$ using \cref{eq:D} \;
    \If{$\exists \ s \neq 1, t: \ d_{s,t} < \gamma$}{
      return $\hat{X}$\;
      }
    $k = k + 1$;
  }
\caption{The pseudo-code of S-ATTack algorithm}
\label{alg:alg}
\end{algorithm}
\end{minipage}
\end{figure}

\section{Experiments}
We conduct the experiments to answer the following questions: 1) How vulnerable are the trajectory prediction models against perturbations on the input sequence?
2) Which of the proposed socially-attended attacks can cause collisions more successfully with smaller perturbations?
3) Are we able to leverage the adversarial examples to improve the robustness of the model? Do they help in better learning the social behavior?

\subsection{Experimental setup}

\subsubsection{Baselines}
In order to show the effectiveness of our attack, we conduct our experiments on six well-established trajectory prediction models. 
\begin{itemize}

\item \textbf{Social-LSTM}
\cite{alahi2016sociallstm} ({\bf S-LSTM}): where a social pooling method is employed to model interactions based on shared hidden states of LSTM trajectory encoders.

\item\textbf{Social-Attention}
\cite{vemula2018socialattention} ({\bf S-Att}): where a self-attention block is in charge of learning interactions between agents.
\item\textbf{Social-GAN}
\cite{gupta2018socialgan} ({\bf S-GAN}): where a max-pooling function is employed to encode neighbourhood information. They leverage a generative adversarial network (GAN) to learn the distribution of trajectories.
\item\textbf{Directional-Pooling}
\cite{kothari2020human} ({\bf D-Pool}): where the features of each trajectory is learned using the relative positions as well as the relative velocity and then pool the learned features to learn social interactions.
\item\textbf{Social-STGCNN}
\cite{mohamed2020stgcnn} ({\bf S-STGCNN}): where graph convolutional neural network is employed to learn the interactions.
\item\textbf{PECNet}
\cite{mangalam2020pecnet} ({\bf PECNet}): where a self-attention based social pooling layer is leveraged with a variational auto-encoder (VAE) network.
\end{itemize}

\subsubsection{Datasets}
\label{sec:dataset}
\noindent\textbf{ETH~\cite{pellegrini2010eth}, UCY~\cite{lerner2007ucy}, and WildTrack~\cite{chavdarova2018wildtrack}}: These are well-established datasets with human positions in world-coordinates. We employ two variants of these datasets for our experiments: (1) for S-LSTM, S-Att, S-GAN and D-Pool baselines, we utilized Trajnet++~\cite{kothari2020human} benchmark which provides identical data splits and data pre-processing. \bl{The observation and prediction lengths are considered as 9 and 12, respectively.}
(2) Since S-STGCNN official code contains its specific pre-processing and data-split on ETH and UCY, we employed the released code to be consistent with their official implementation. \bl{Here, the observation and prediction lengths are considered as 8 and 12, respectively.}  

\noindent\textbf{SDD~\cite{robicquet2016learning}}: The Stanford Drone Dataset is a human trajectory prediction dataset in bird's eye view. PECNet is one of the state-of-the-art methods with official published code on this dataset. Hence, we report PECNet performance on this dataset. The observation and prediction lengths are considered as 8 and 12, respectively. 

\subsubsection{Implementation details}
We set the maximum number of iterations to $100$.
Inspired by the localization algorithm errors mentioned in the introduction, the maximum size of the perturbation for each point $\epsilon$ is considered $0.2$~m. Also, we set $\lambda_r$ and $\lambda_w$ in \cref{eq:hard} and \cref{eq:soft} equal to $0.1$ and $0.5$ respectively. 
The full list of hyperparameters will be provided in the supplementary material.

\begin{figure*}[ht]
  \centering
\begin{subfigure}[b]{0.327\linewidth}
    \centering\includegraphics[width=\linewidth]{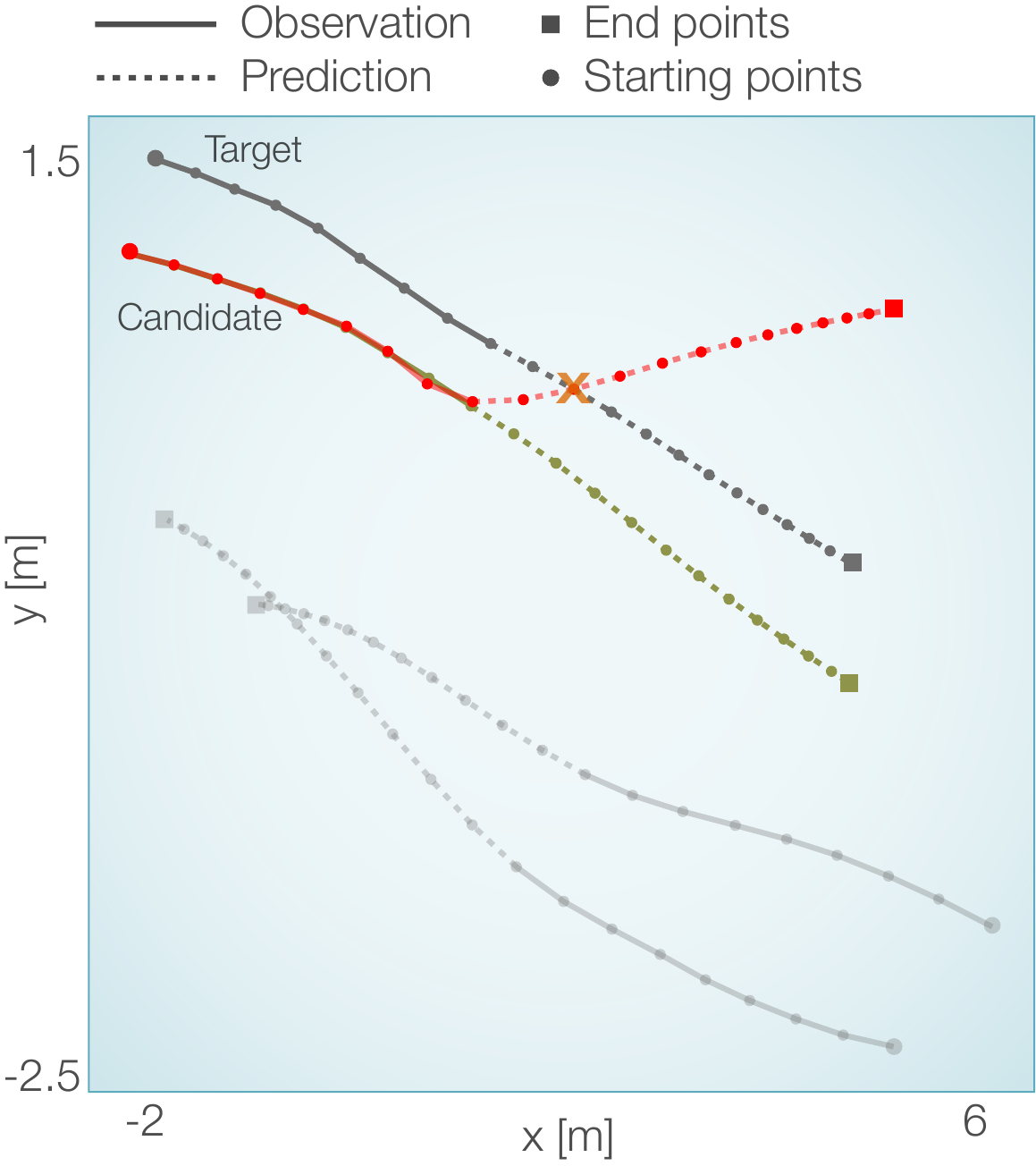}
    \caption{~S-LSTM (P-avg : $0.009$m)}
    \label{fig:all_methods_slstm}
  \end{subfigure}
  \hfill
    \begin{subfigure}[b]{0.327\linewidth}
    \centering\includegraphics[width=\linewidth]{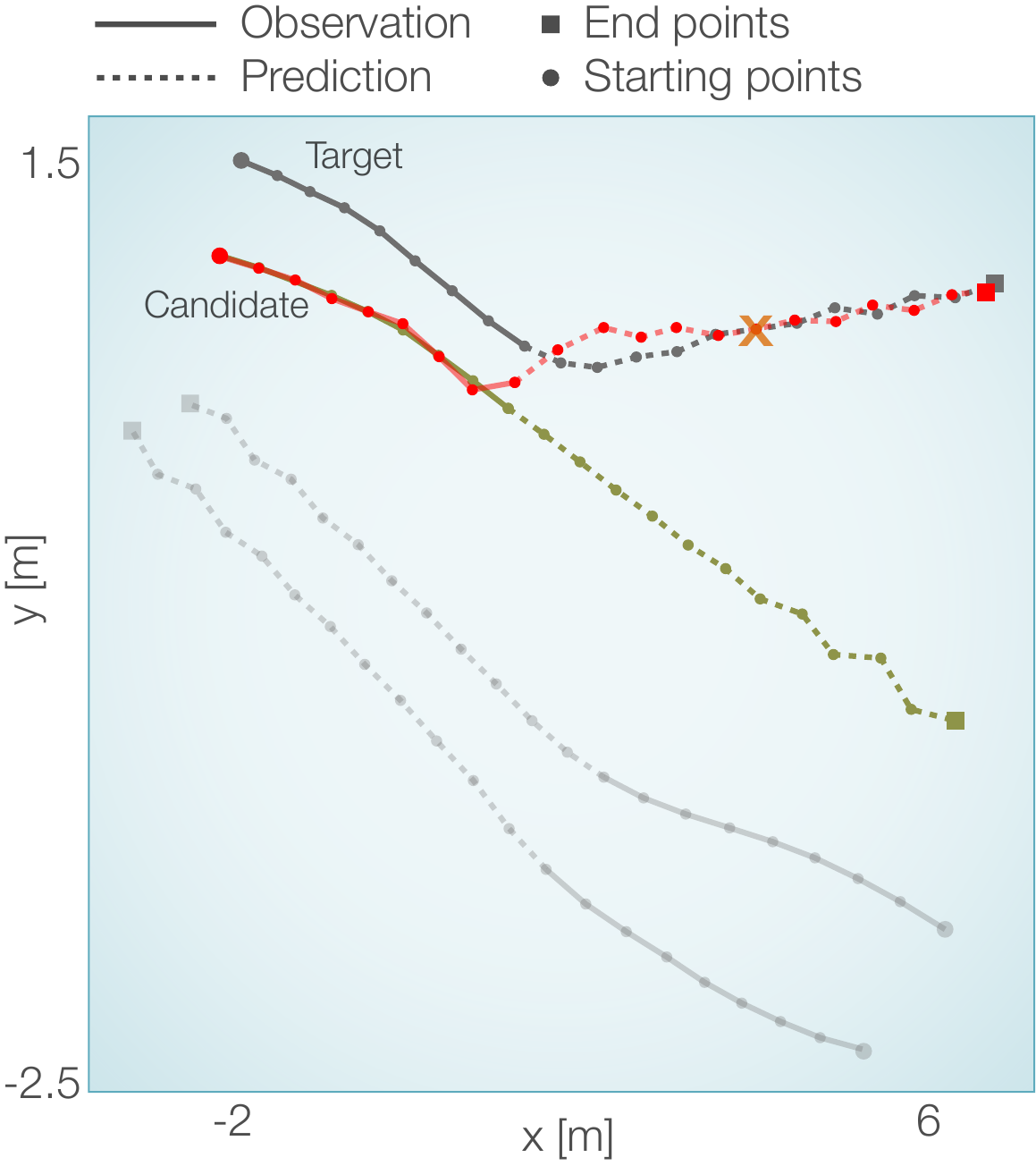}
    \caption{~S-Att, (P-avg : $0.028$m)}
    \label{fig:all_methods_satt}
  \end{subfigure}
  \hfill
  \begin{subfigure}[b]{0.327\linewidth}
    \centering\includegraphics[width=\linewidth]{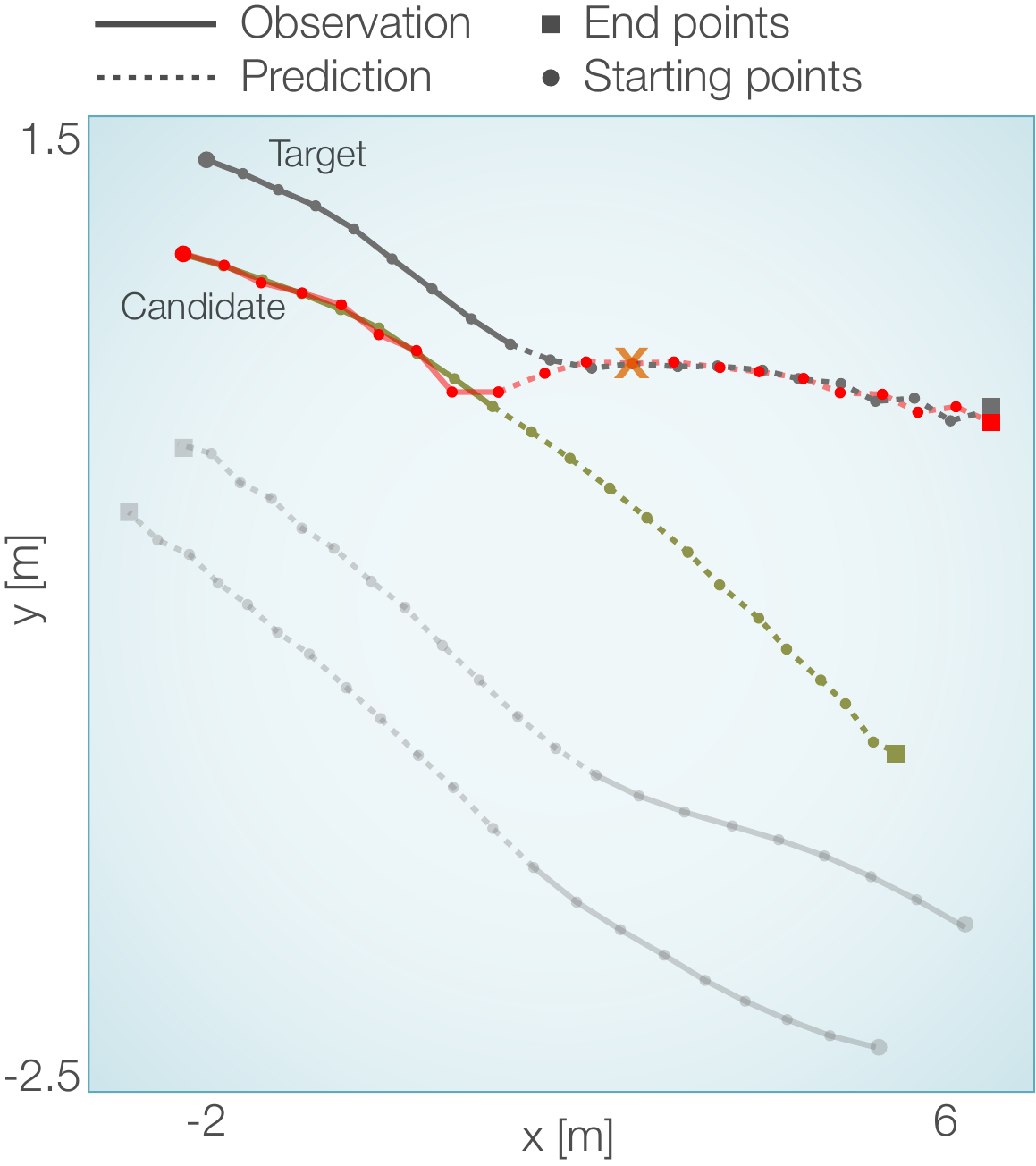}
    \caption{~D-Pool, (P-avg : $0.029$m)}
    \label{fig:all_methods_dpool}
  \end{subfigure}
  
  \begin{subfigure}[b]{0.327\linewidth}
    \centering\includegraphics[width=\linewidth]{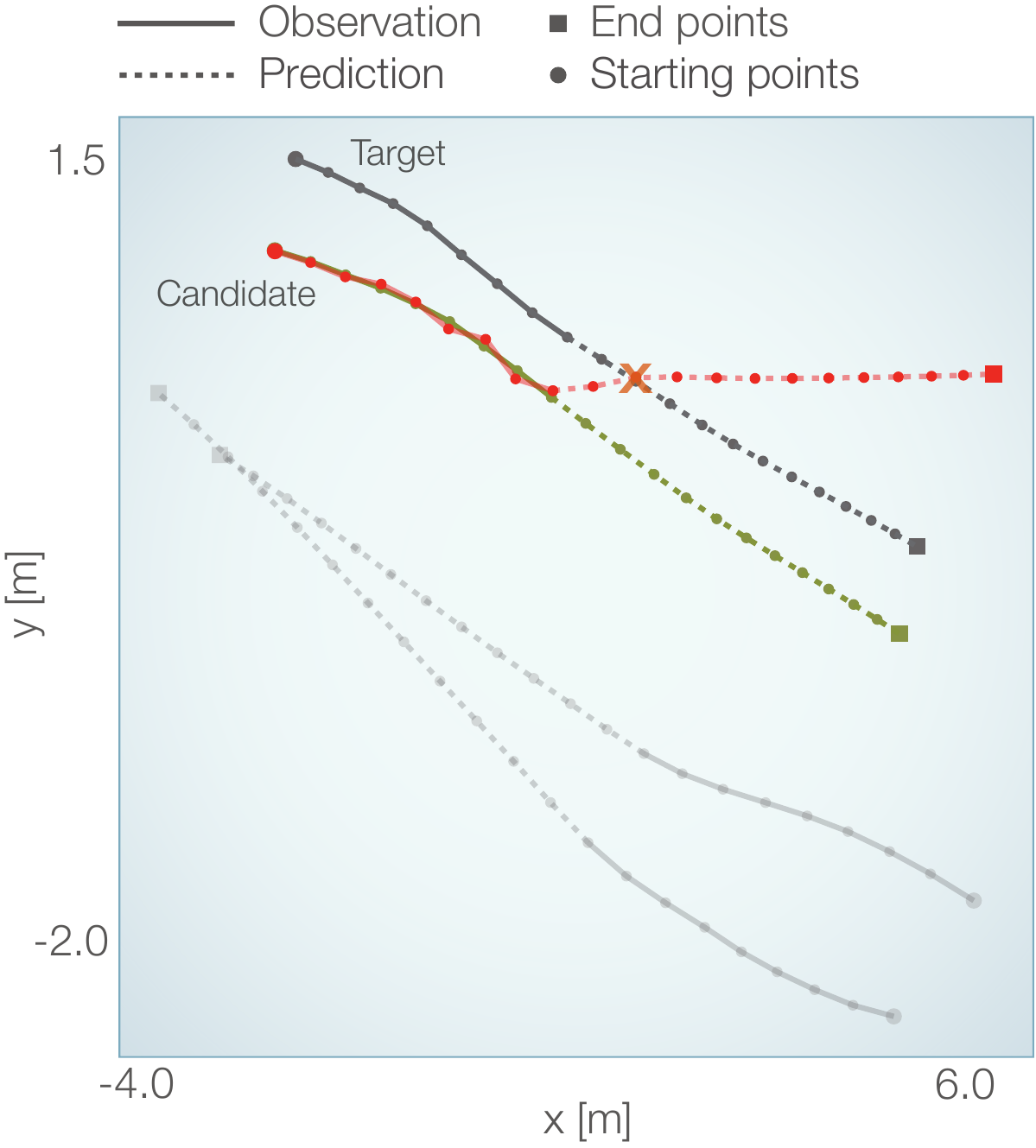}
    \caption{~S-GAN, (P-avg : $0.022$m)}
    \label{fig:all_methods_sgan}
  \end{subfigure}%
    \begin{subfigure}[b]{0.327\linewidth}
    \centering\includegraphics[width=\linewidth]{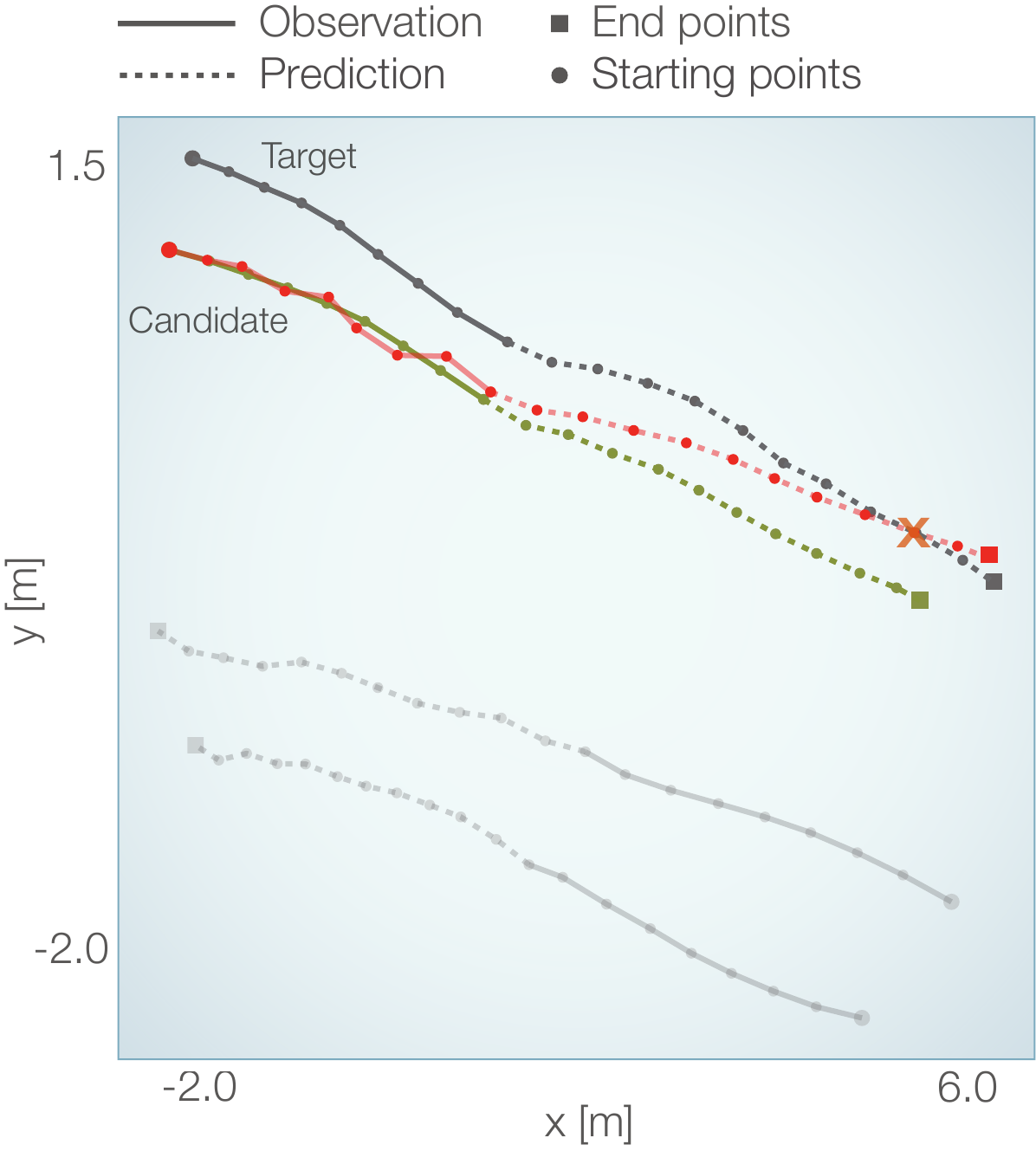}
    \caption{~S-STGCNN, (P-avg : $0.042$m)}
    \label{fig:all_methods_sstgcnn}
  \end{subfigure}%
  \caption{
  Comparison of the performance of different models under our attack. The candidate agent is depicted with green before the perturbation and red after it. For brevity, we have not shown the prediction of neighboring agents before the attack. The scale of y axis is enlarged to better show the difference. The orange X denotes the target point. Our attack achieves collisions with adding small perturbations.
}
 \label{fig:all_methods}
\end{figure*}

\subsubsection{Metrics}
In the experiments, we report the performances according to the following metrics: 

\begin{itemize} 
    \item Collision Rate (\textbf{CR}): this metric measures the existence of collision in the predicted trajectories of the model.
    Indeed, it calculates the percentage of samples in which at least one collision in the predicted trajectories between the candidate agent and its neighbors occurs. This metric assesses whether the model learned the notion of collision avoidance.
    Note that we set the distance threshold for indicating a collision $\gamma$ in \cref{eq:formulation2} equal to $0.2$~m .

    \item Perturbation average (\textbf{P-avg}): 
    the average of perturbation sizes at each timestep which is added to the input observation. The numbers are reported in meters.

    \item Average / Final Displacement Error (\textbf{ADE/FDE}): the average/final displacement error between the predictions of the model and the ground-truth values. This metric is commonly used to report the performance of trajectory prediction models and is reported in meters.
\end{itemize}

\subsection{Attack results}
\label{sec:results}

We first provide the quantitative results of applying S-ATTack to the baselines in~\Cref{tab:all_methods}. The results indicate a substantial increase in the collision rate (at least $3$ times) across all baselines by adding perturbations with P-avg smaller than $0.11$~m. This questions the social behavior of the models in terms of collision avoidance. 

\Cref{fig:all_methods} visualizes the performance of the baselines S-LSTM, S-Att, D-Pool, S-GAN and S-STGCNN under our attack with the same input. Note that all the baselines are trained on the first group of datasets in \ref{sec:dataset}.
S-LSTM does not change its predictions after adding perturbations but S-Att and D-Pool counteract to avoid collisions. This shows that there exists some collision avoidance behavior understanding in some prediction models, but they are not enough for avoiding a collision.
We show more successful cases and also some failure cases in the supplementary material.

\begin{table}
\begin{center}
\begin{tabular}{|l|c|c|c|}
\hline
\multirow{2}{4em}{Model} & Original & \multicolumn{2}{c|}{Attacked} \\
& CR~[$\%$]  $\downarrow$ & CR~[$\%$] $\downarrow$& P-avg~[m] $\downarrow$\\
\hline\hline
S-LSTM \cite{alahi2016sociallstm} & $7.8$ & $89.8$ & $0.031$ \\
S-Att \cite{vemula2018socialattention} & $9.4$ & $86.4$ & $0.057$\\
S-GAN \cite{gupta2018socialgan} & $13.9$ & $85.0$ & $0.034$ \\
D-Pool \cite{kothari2020human} & $7.3$ & $88.0$ & $0.042$\\
\hline
S-STGCNN \cite{mohamed2020stgcnn} & $16.3$ & $59.1$ & $0.11$ \\
\hline
PECNet \cite{mangalam2020pecnet} & $15.0$ & $64.9$ & $0.071$ \\
\hline
\end{tabular}
\end{center}
\caption{Comparing the performance of different baselines before (Original) and after the attack (Attacked). Horizontal lines separate models with different datasets.}
\label{tab:all_methods}
\end{table}

As D-pool performs better than others in terms of collision avoidance before attack, in the rest of the paper, we conduct our main experiments on it.

\subsection{Comparison of different attention methods}
\label{sec:ablation}

\begin{table}[!t]
\begin{center}
\begin{tabular}{|l|c|c|}
\hline
Attacks & CR~[$\%$] $\downarrow$ & P-avg~[m] $\downarrow$ \\
\hline\hline
Random noise  & $17.6$ & $0.199$\\
No-attention  \cref{eq:naive} & $44.7$ & $0.179$\\
Hard-attention  \cref{eq:hard} & $84.8$ & $0.041$ \\
Soft-attention  \cref{eq:soft} & \textbf{$88.0$} & \textbf{$0.042$} \\
\hline
\end{tabular}
\end{center}
\caption{Comparing different proposed attack methods on D-Pool. }
\label{tab:ablation}
\end{table}

\begin{figure*}[t]
  \centering
\begin{subfigure}[b]{0.327\linewidth}
    \centering\includegraphics[width=\linewidth]{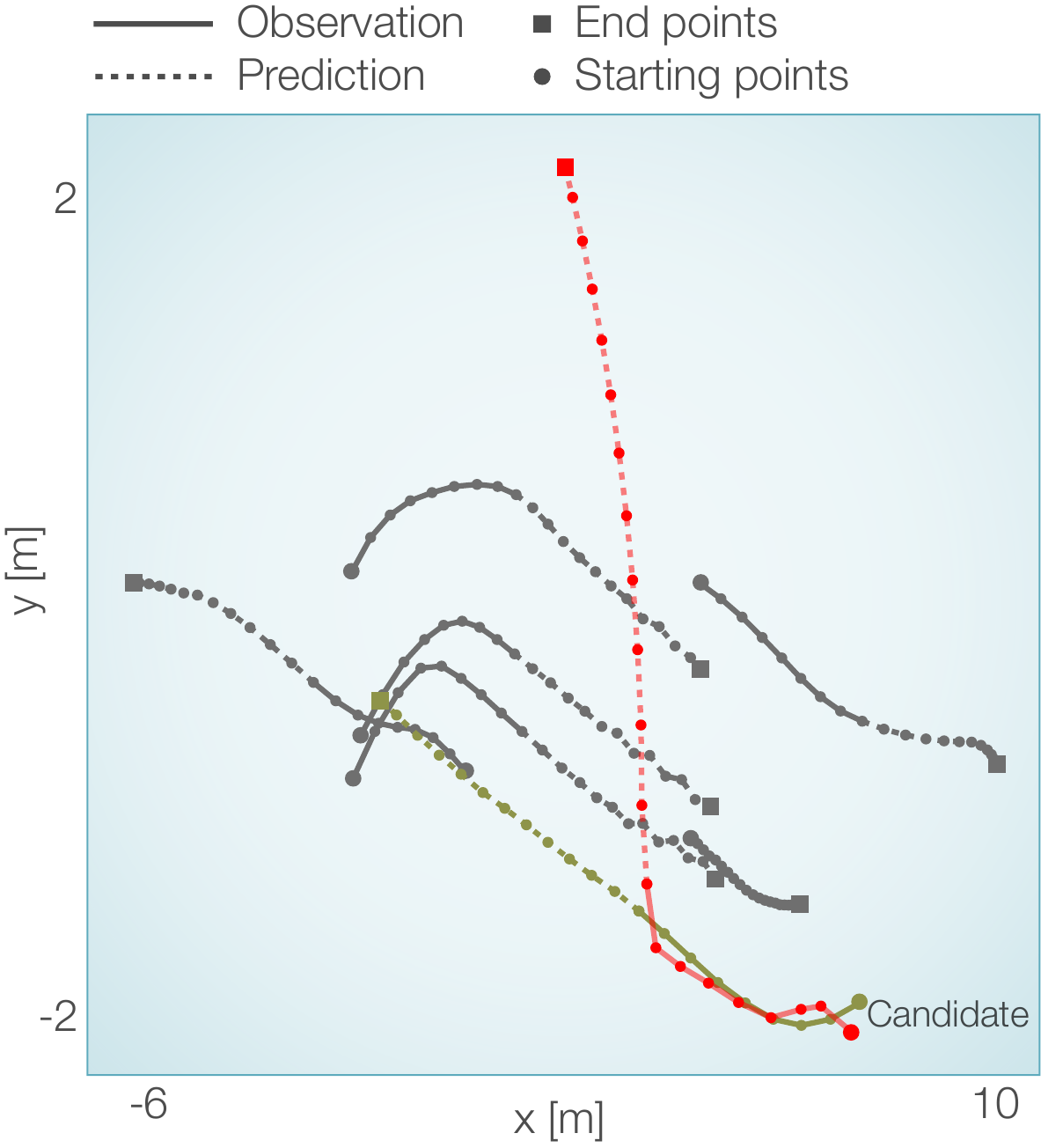}
    \caption{No-attention (P-avg : $0.155$m)}
  \end{subfigure}
  \hfill
    \begin{subfigure}[b]{0.327\linewidth}
    \centering\includegraphics[width=\linewidth]{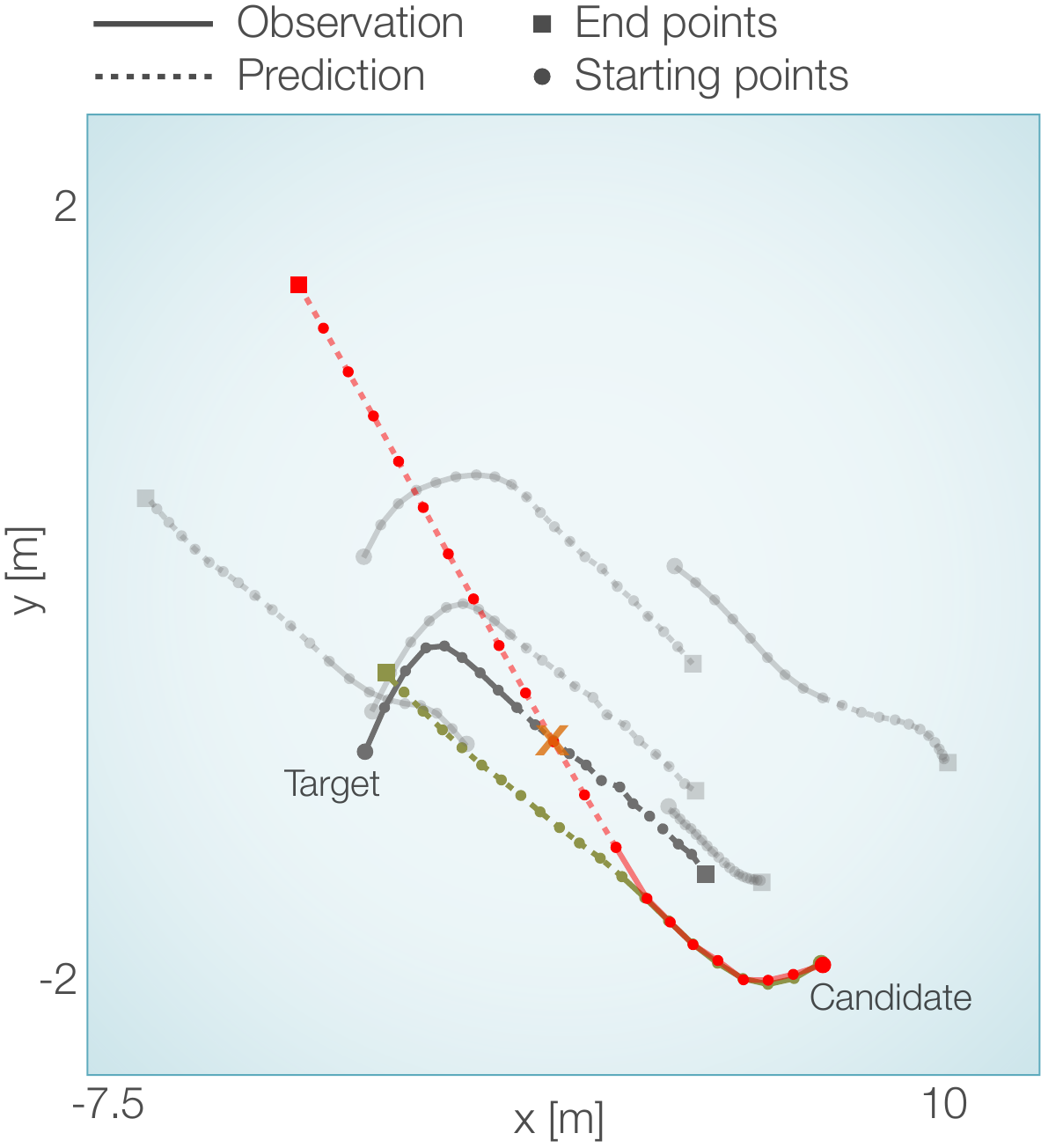}
    \caption{Hard-attention (P-avg : $0.04$m)}
  \end{subfigure}
  \hfill
  \begin{subfigure}[b]{0.327\linewidth}
    \centering\includegraphics[width=\linewidth]{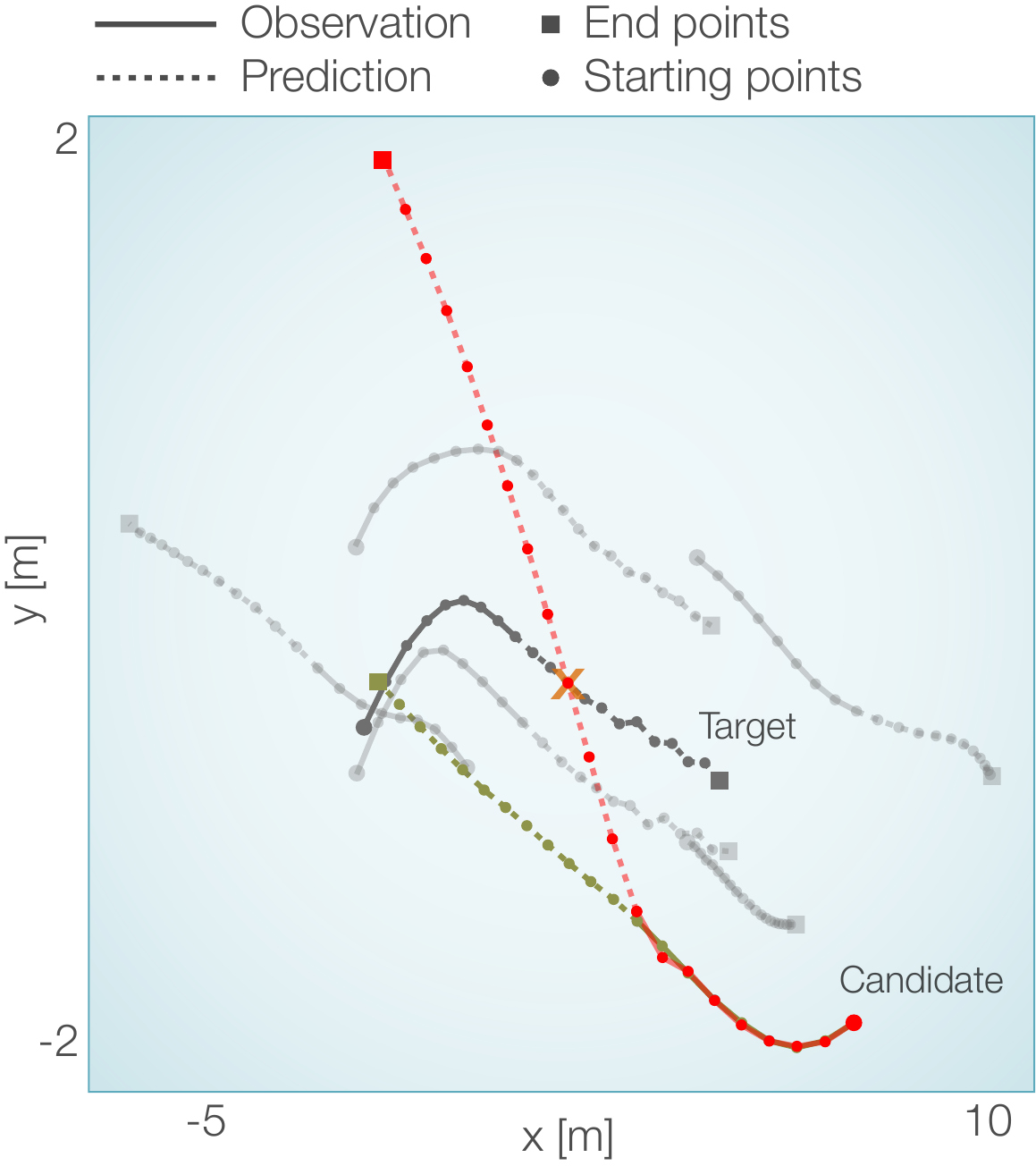}
    \caption{Soft-attention (P-avg : $0.014$m)}
  \end{subfigure}%
  \caption{
  Comparison of different attack approaches. The no-attention approach \cref{eq:naive} could not make a collision despite large P-avg. The hard-attention approach \cref{eq:hard} led to a collision with a smaller P-avg. A better target point for collision is found by soft-attention \cref{eq:soft} leading to a collision with the least P-avg.}
 \label{fig:ablation}
\end{figure*}

In this section, we compare how different strategies for choosing the collision point affects the collision rate and the perturbation size.
The quantitative results are shown in \Cref{tab:ablation}. 
We report the results of \bl{ Gaussian random noise with variance of 0.2}, no-attention \cref{eq:naive}, hard-attention \cref{eq:hard}, and soft-attention \cref{eq:soft}. Random noise and no-attention approaches have small collision rates with large perturbation sizes which indicates the need for a strategy for selecting the collision point.
Both hard-attention and soft-attention mechanisms substantially improve collision rate. However, the freedom in the soft-attention approach lets it smartly find better collision points in some of the samples leading to a higher collision rate. \Cref{fig:ablation} visualizes the performance of different attack approaches for one data sample. In the illustrated example, in contrast to other two approaches, Soft-attention targets a collision point which is further but easier to collide, thus, leads to a smaller perturbation size.

\subsection{Transferability}
\label{sec:transfer}
A common practice in adversarial attack studies is to investigate the transferability of perturbations achieved for one model across other models. This shows the existence of common weaknesses across different models or biases in the dataset leading to brittle features in the models. Note that due to the inconsistency of the data used for different baselines, we perform this study only for the models which use the Trajnet++ data.

\begin{figure*}[ht]
  \centering
  \begin{subfigure}[b]{0.327\linewidth}
    \centering\includegraphics[width=\linewidth]{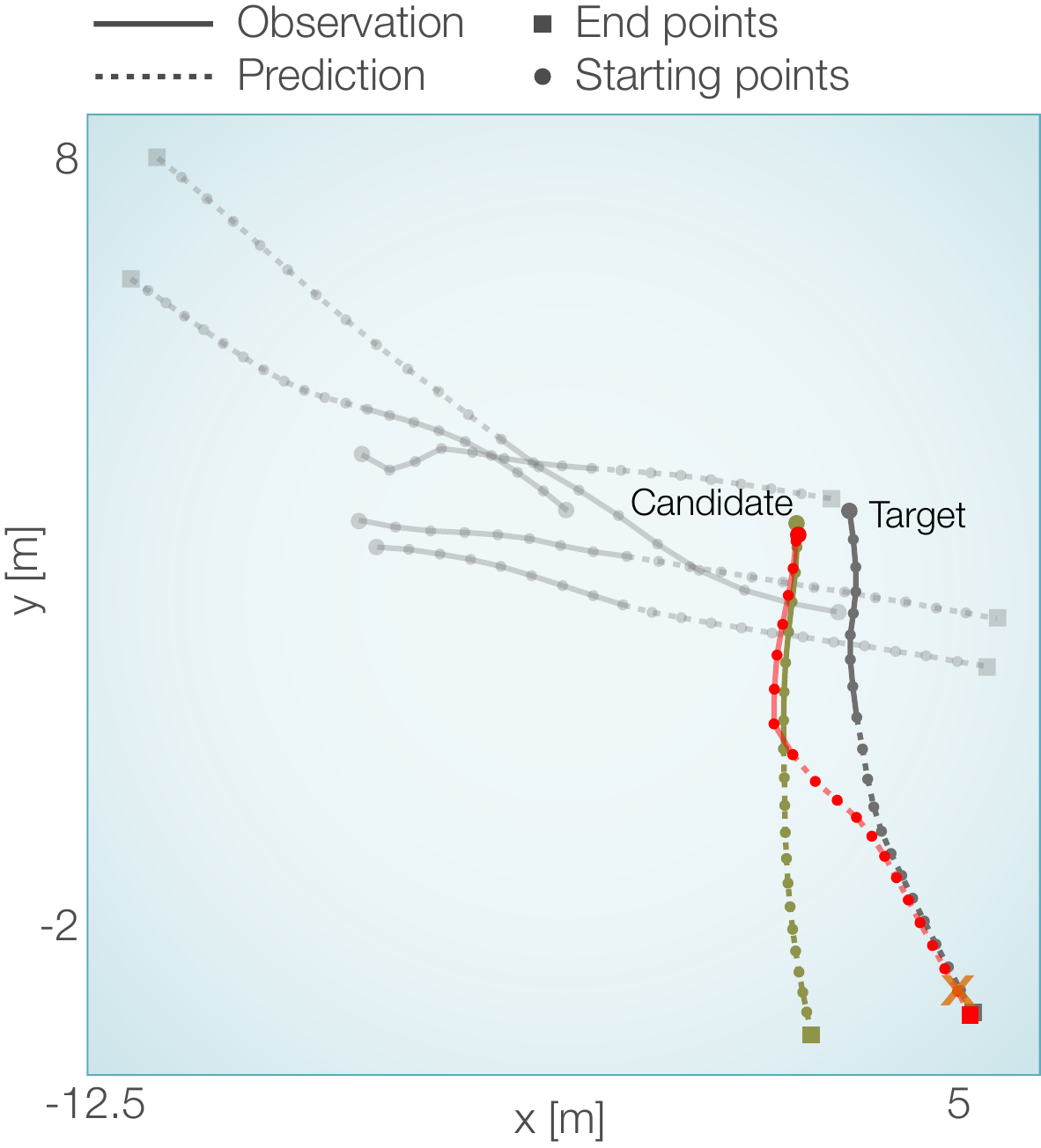}
    \caption{~D-Pool}
  \end{subfigure}%
  \hfill
    \begin{subfigure}[b]{0.327\linewidth}
    \centering\includegraphics[width=\linewidth]{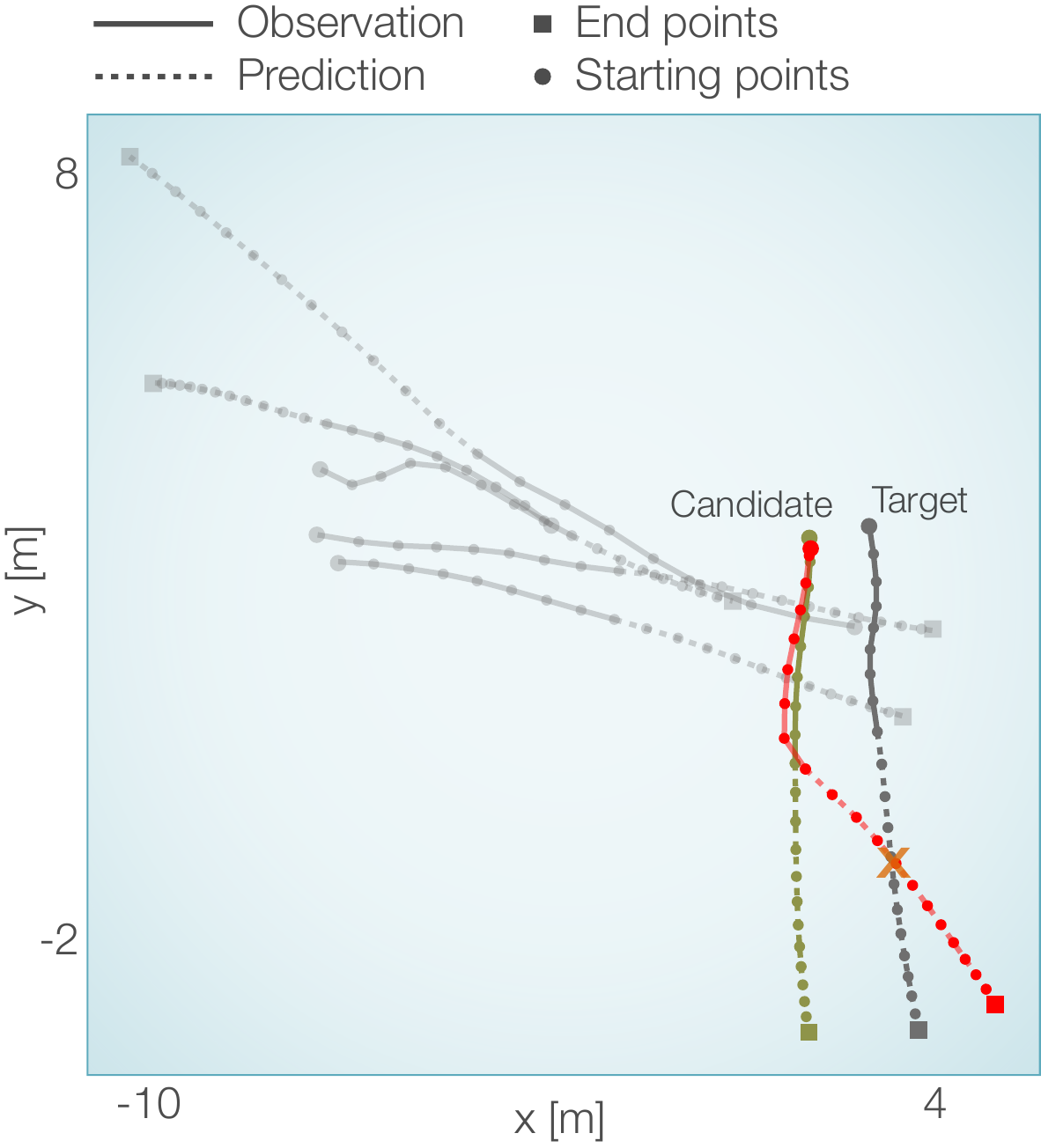}
    \caption{~S-LSTM}
  \end{subfigure}
  \hfill
    \begin{subfigure}[b]{0.327\linewidth}
    \centering\includegraphics[width=\linewidth]{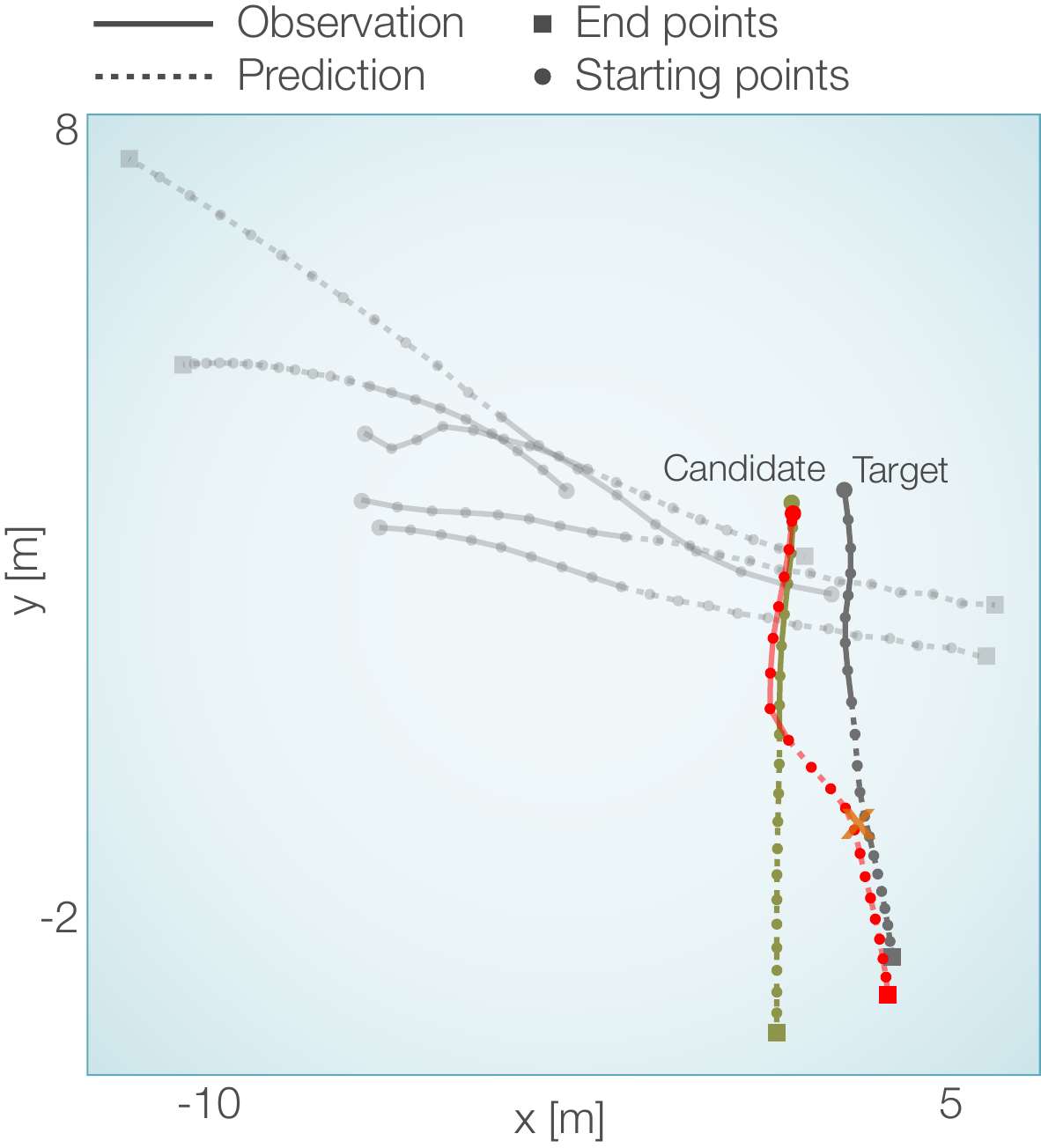}
    \caption{~S-Att}
  \end{subfigure}
  \caption{
  Transferring an adversarial example obtained from attacking D-Pool model to  S-LSTM and S-Att models. P-avg is $0.151$~m. We can observe that the perturbation generates collisions in both S-LSTM and S-Att models, although not optimized for them.}
 \label{fig:transfer}
\end{figure*}

To study how the generated perturbations can transfer to other models, source models were attacked and the achieved perturbations were used to evaluate others as target models. In addition to the data-driven models, we study the performance of a physics-based model, Social-forces (S-Forces) \cite{helbing1998social-forces} against generated perturbations. Note that S-Forces has zero collision rate before the attack.
\Cref{tab:transferability} shows that the transferred perturbations can make substantial collision rates for the data-driven models. This means that there exist common defects across different models. Expectedly, S-Forces is robust against the perturbations. This is due to the fact that collision avoidance is an explicit rule defined for the model. However, this robustness comes with the cost of accuracy loss as reported in \cite{alahi2016sociallstm}. We will further analyze our findings in \cref{sec:discussions}. 
\Cref{fig:transfer} shows the result of one identical perturbation on three different models.

\begin{table}[!t]
\begin{center}

\begin{tabular}{|l|c|c|c|c|}
\hline
Target models & \multicolumn{4}{c|}{Source models} \\
\hline\hline
& S-LSTM & S-Att & S-GAN & D-Pool\\
\hline

S-LSTM \cite{alahi2016sociallstm} &89.8 &82.7&  85.1 & 68.3 \\
S-Att  \cite{vemula2018socialattention} &53.0 &86.4& 57.8 & 70.0 \\
S-GAN \cite{gupta2018socialgan} &40.8 &59.7& 85.0 &84.1  \\
D-Pool \cite{kothari2020human}  &88.4  &81.3& 55.6 & 88.0\\
S-Forces \cite{helbing1998social-forces} &0.69 &0.70& 1.30 & 0.60  \\
\hline
\end{tabular}
\end{center}
\caption{Studying the transferability of adversarial examples. The adversarial examples are learned for source models and are transferred to the target models for the evaluation. The reported numbers are Collision Rate (CR) values.}

\label{tab:transferability}
\end{table}

\subsection{Enhancing the social understanding}
\label{sec:robust}
We utilize our S-ATTack to improve the collision avoidance of the model. To this end, we employ a similar approach to~\cite{madry2017towards}. We fine-tune the model using a combination of the original training data and the adversarial examples generated by our S-ATTack method.
In this experiment, we set the maximum perturbation size $\epsilon$ equal to $0.03$.

\Cref{tab:robustness} indicates that the model's collision avoidance could improve by $11\%$. Moreover, the collision rate after attack improves by $60\%$ meaning that it is much less vulnerable to the attack. As shown in the table, fine-tuning the model with random noise could not improve the collision avoidance.
Therefore, we conclude that our adversarial examples provide useful information to improve the collision avoidance of the model.
Note that the prediction error of the model in terms of ADE/FDE is slightly increased. This shows a trade-off between accuracy and robustness. This can be similar to the findings in the previous works on image classifiers \cite{tsipras2018robustness}. 

\begin{table*}[!t]
\begin{center}
\begin{tabular}{|l|c|c c||c c|}
\hline
& & \multicolumn{2}{c||}{Original} & \multicolumn{2}{c|}{Attacked}  \\
& ADE/FDE [m] $\downarrow$ & CR [$\%$] $\downarrow$ & CR gain [$\%$] $\uparrow$ & CR [$\%$] $\downarrow$ & CR gain [$\%$]  $\uparrow$  \\
\hline\hline
D-Pool & 0.57 / 1.23 & 7.3 & - & 37.3 & - \\
D-Pool w/ rand noise & 0.57 / 1.23 & 7.5 & -2.7 & 36.1 & +3.2\\
D-Pool w/ S-ATTack & 0.60 / 1.28 & \textbf{6.5} & \textbf{+10.4} & \textbf{14.7} & \textbf{+60}\\
\hline
\end{tabular}
\end{center}
\caption{Comparing the original model and the fine-tuned model with random-noise data augmentation (D-Pool  w/ rand noise) and S-ATTack adversarial examples (D-Pool w/ S-ATTack). The numbers are on Trajnet++ challenge testset. ADE, FDE are reported in meters.}
\label{tab:robustness}
\end{table*}

\begin{figure}[ht]
  \centering
\begin{subfigure}[b]{0.4\linewidth}
    \centering\includegraphics[width=150pt]{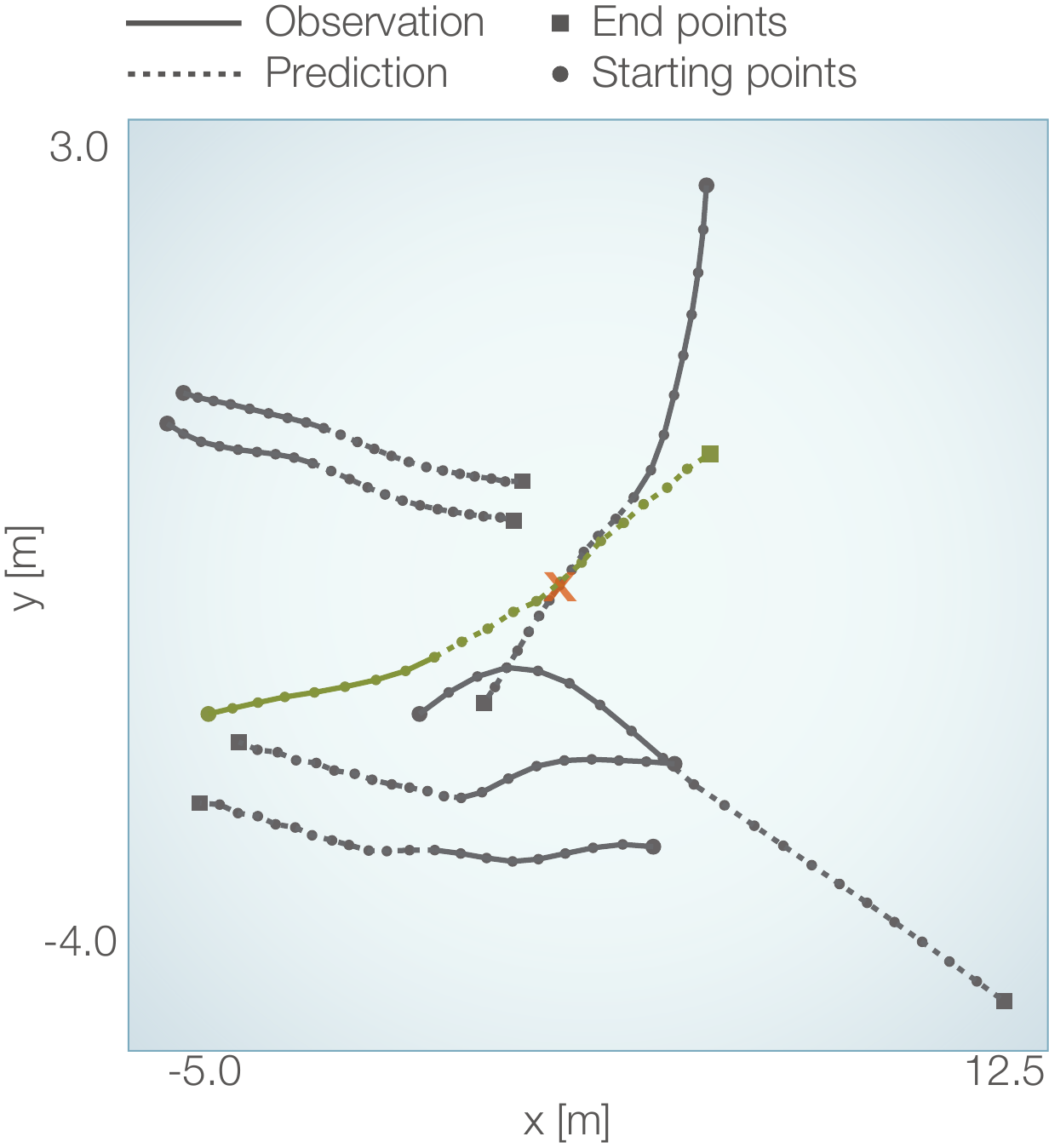}
    \caption{~D-Pool \\ Original}
  \end{subfigure}
    \begin{subfigure}[b]{0.4\linewidth}
    \centering\includegraphics[width=150pt]{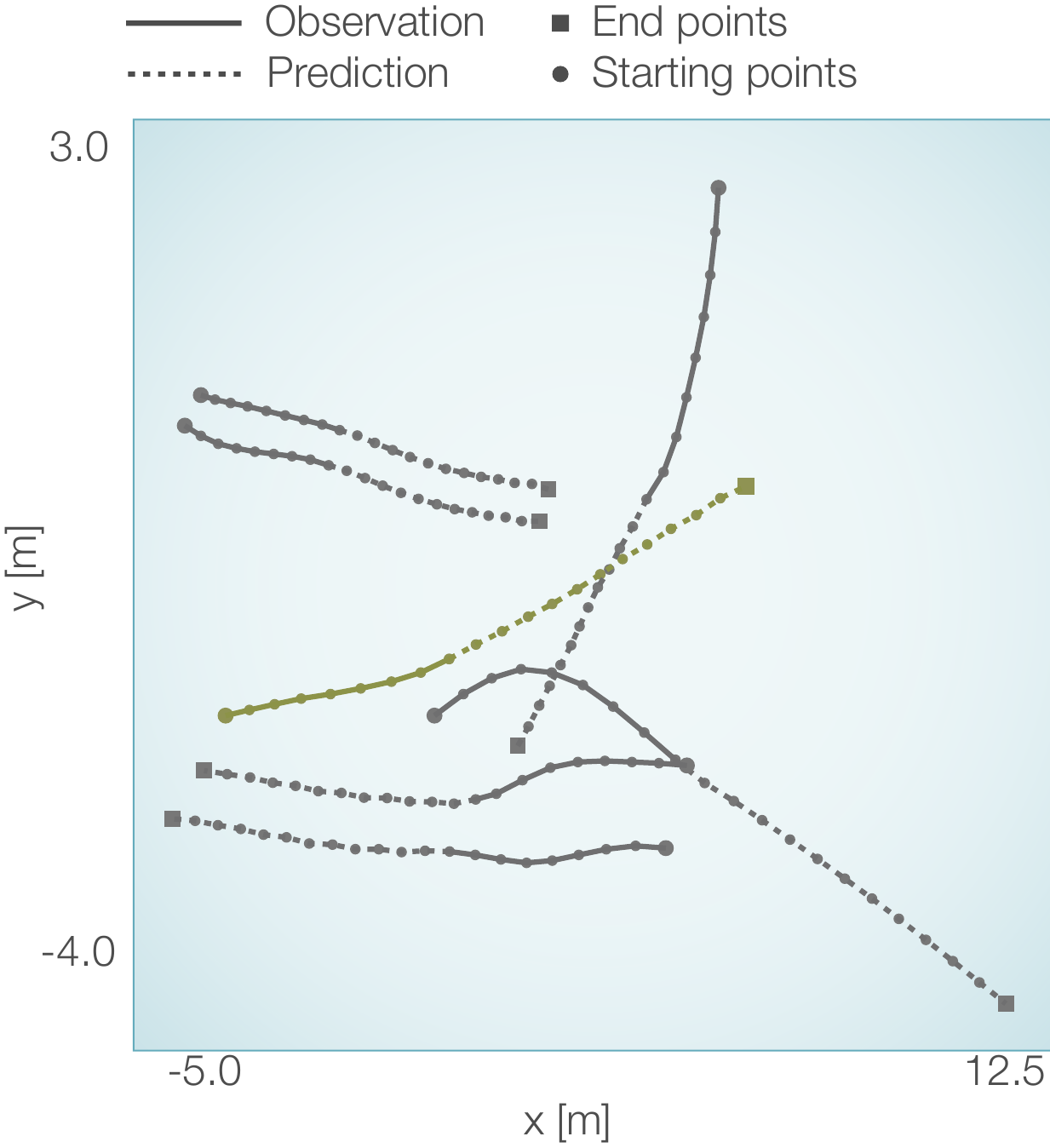}
    \caption{~D-Pool w/ S-ATTack \\ Original}
  \end{subfigure}
  \caption{
  Comparison of the performance of the original model and the enhanced one with S-ATTack. The original model has a collision in its predictions (a) while the enhanced model prediction is collision-free (b).}
 \label{fig:robust}
\end{figure}

We also show the performance of these models visually. In \Cref{fig:robust} (a), the original model's prediction with an occurring collision is seen while the enhanced model prediction is collision-free (\Cref{fig:robust} (b)). In the next figure, \Cref{fig:robust2} (a), we can observe that attacking D-Pool leads to a collision with a P-avg of $6.6$ but the enhanced model is robust against attack (\Cref{fig:robust2} (b)) indicating a better collision avoidance understanding. 

\begin{figure}[ht]
  \centering
  \begin{subfigure}[b]{0.4\linewidth}
    \centering\includegraphics[width=150pt]{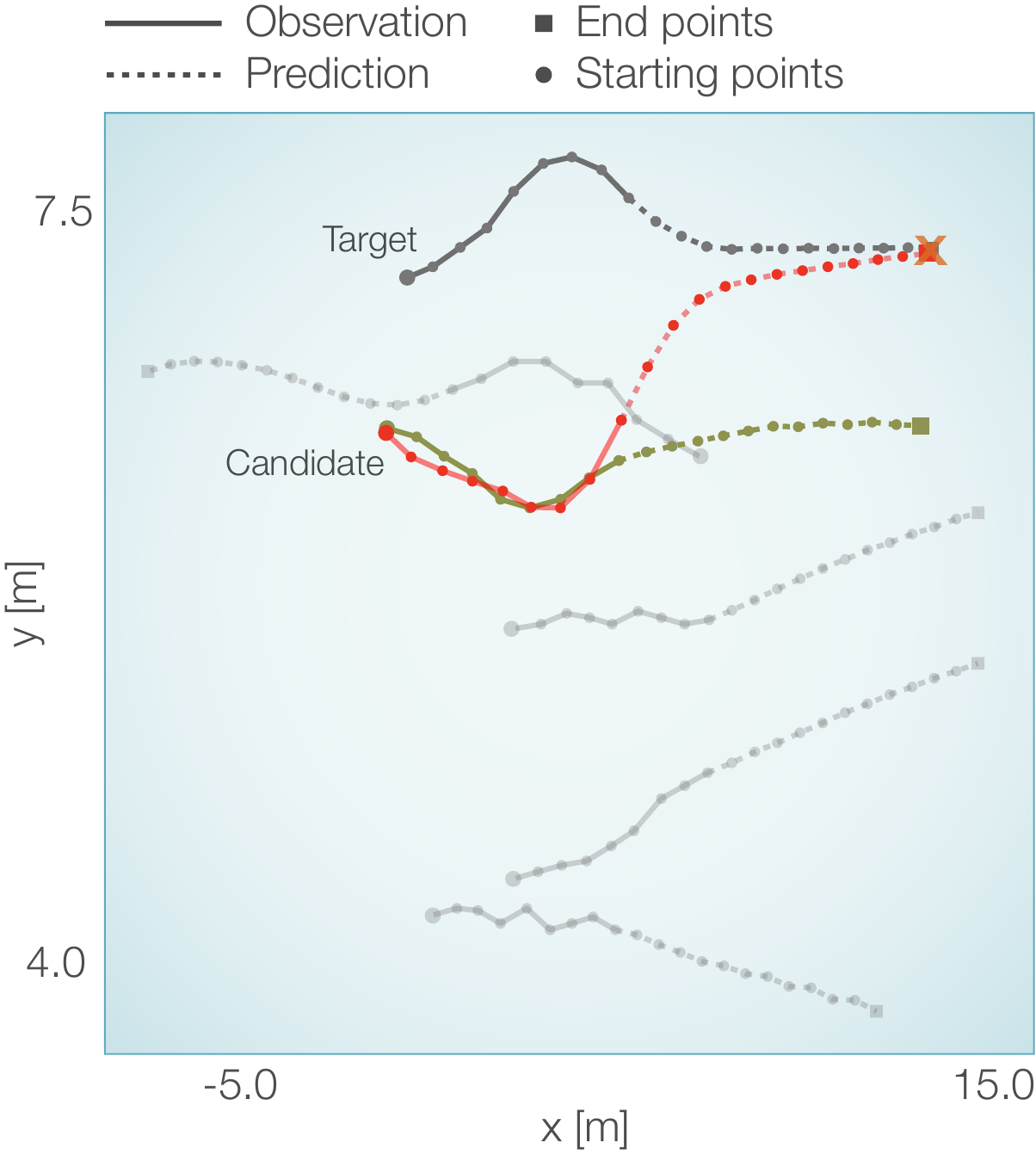}
    \caption{~D-Pool \\ Attacked}
  \end{subfigure}
  \begin{subfigure}[b]{0.4\linewidth}
    \centering\includegraphics[width=150pt]{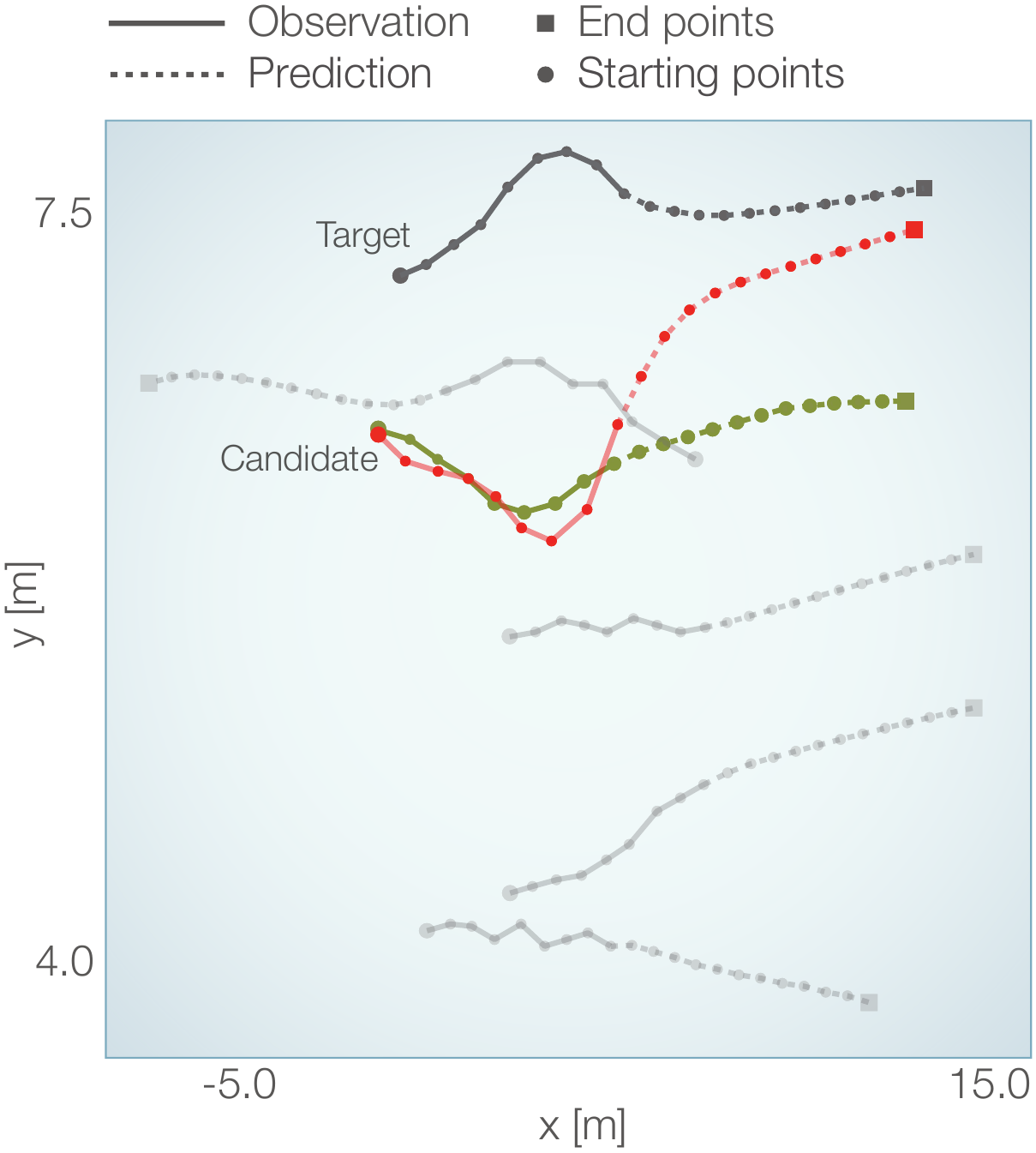}
    \caption{~D-Pool w/ S-ATTack \\ Attacked}
  \end{subfigure}
  \caption{
  Comparison of the vulnerability of the victim model under attack and the enhanced one with S-ATTack. While D-Pool can be attacked to cause a collision with a P-avg of $6.6$ (a), the enhanced model cannot be attacked even with a large perturbation (b). }
 \label{fig:robust2}
\end{figure}

\subsection{Discussions}
\label{sec:discussions}
In this section, equipped with the performed experiments in previous parts, we want to shed light on the weaknesses of the studied trajectory prediction models. Also, we will mention the limitations of our work.

\begin{enumerate}

\item Although the trajectory prediction models are designed to capture interactions among people, our results in~\cref{sec:results} showed that they are fragile and cannot generalize to perturbed data. Moreover, we showed in~\cref{sec:transfer} that the perturbations are transferable across different models indicating the existence of common weaknesses across these models.
We have two hypotheses as the reasons of high vulnerability to the attack and transferability of the perturbations: 1) this can be due to the lack of collision avoidance inductive bias in the models. While inductive bias is more influential when the size of the training data is limited, it may be of less impact in large data regime. 2) because of limited training data, there exists unexplored input space for the predictor model. We leave a detailed study on these points for future work.

\item The success of our attack in making collisions may raise the concern that the models are actually unaware of the social interactions. We perform an experiment to assess this point. We attack the model while keeping the predictions of neighboring agents frozen for the attack algorithm. Therefore, the attack cannot consider other agents' counteracts to the perturbations. As shown in~\Cref{fig:justego}, the attack fails in causing collision. This result indicates that while S-ATTack can cause collisions in the models' predictions, the models have limited social understanding allowing them to counteract perturbations. 

\begin{figure}[!t]
\begin{center}
\includegraphics[width=0.4\linewidth]{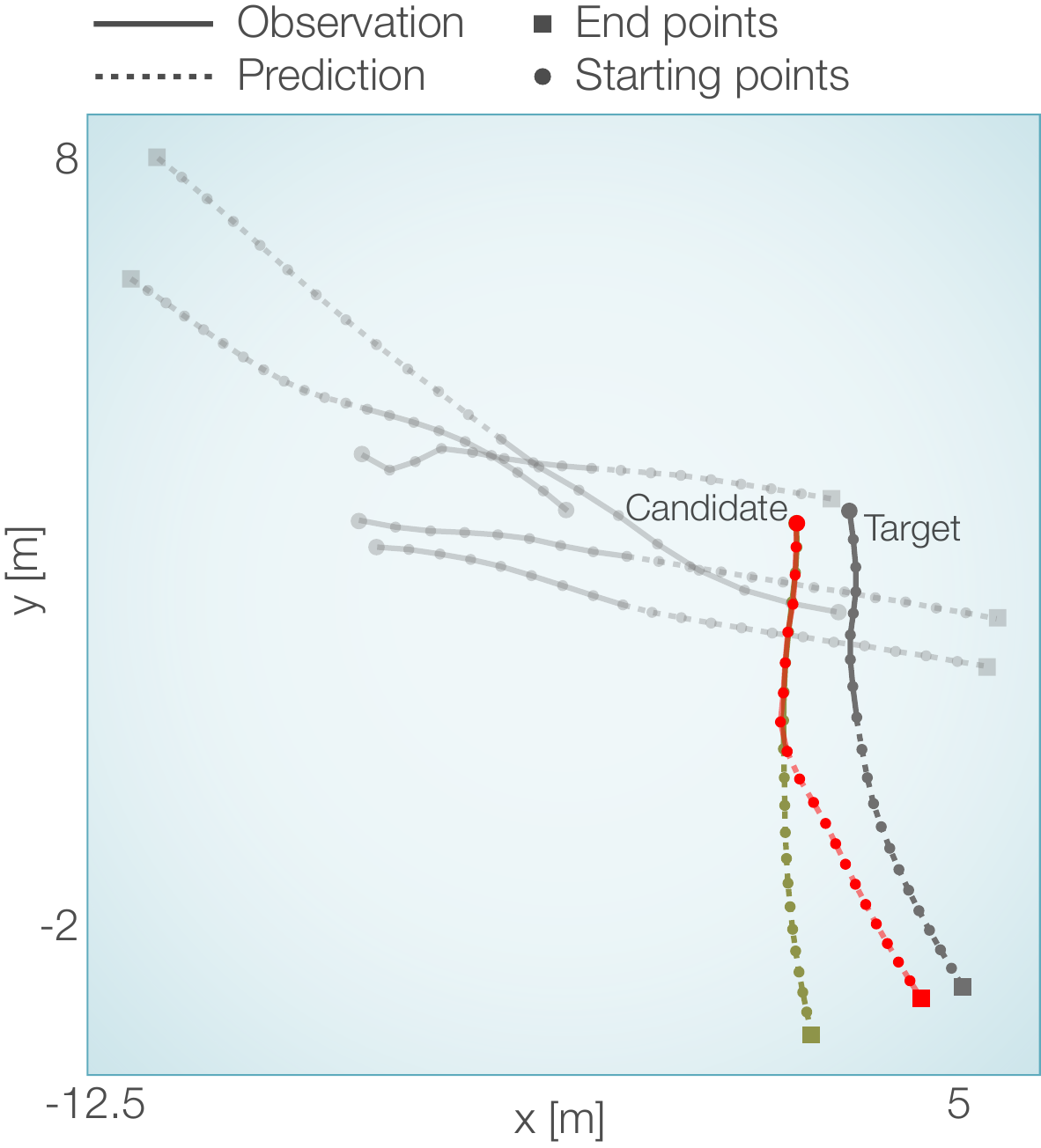}
\end{center}
  \caption{We froze the predictions of neighboring agents for the attack algorithm. The algorithm fails in making collision since it is unaware of the counteracts of neighbors to the perturbations. This indicates that the prediction model counteracts with the perturbations. Also, the attack algorithm requires to consider other agent's counteracts to cause a collision. The same sample was successfully attacked in \Cref{fig:transfer}.
  }
\label{fig:justego}
\end{figure}

\item To compare the models' sensitivity to the perturbations added to different observation points, we add small random noise (less than $0.2$m) to each point separately and measure its effect on the prediction error.
\Cref{fig:histogram} shows that the predictors are over-dependant on the last observation point. It is consistent with our findings that the perturbations generated by S-ATTack tend to have larger components on the last timesteps trying to use this feature. 

\begin{figure}[!t]
\begin{center}
    \includegraphics[width=0.65\columnwidth]{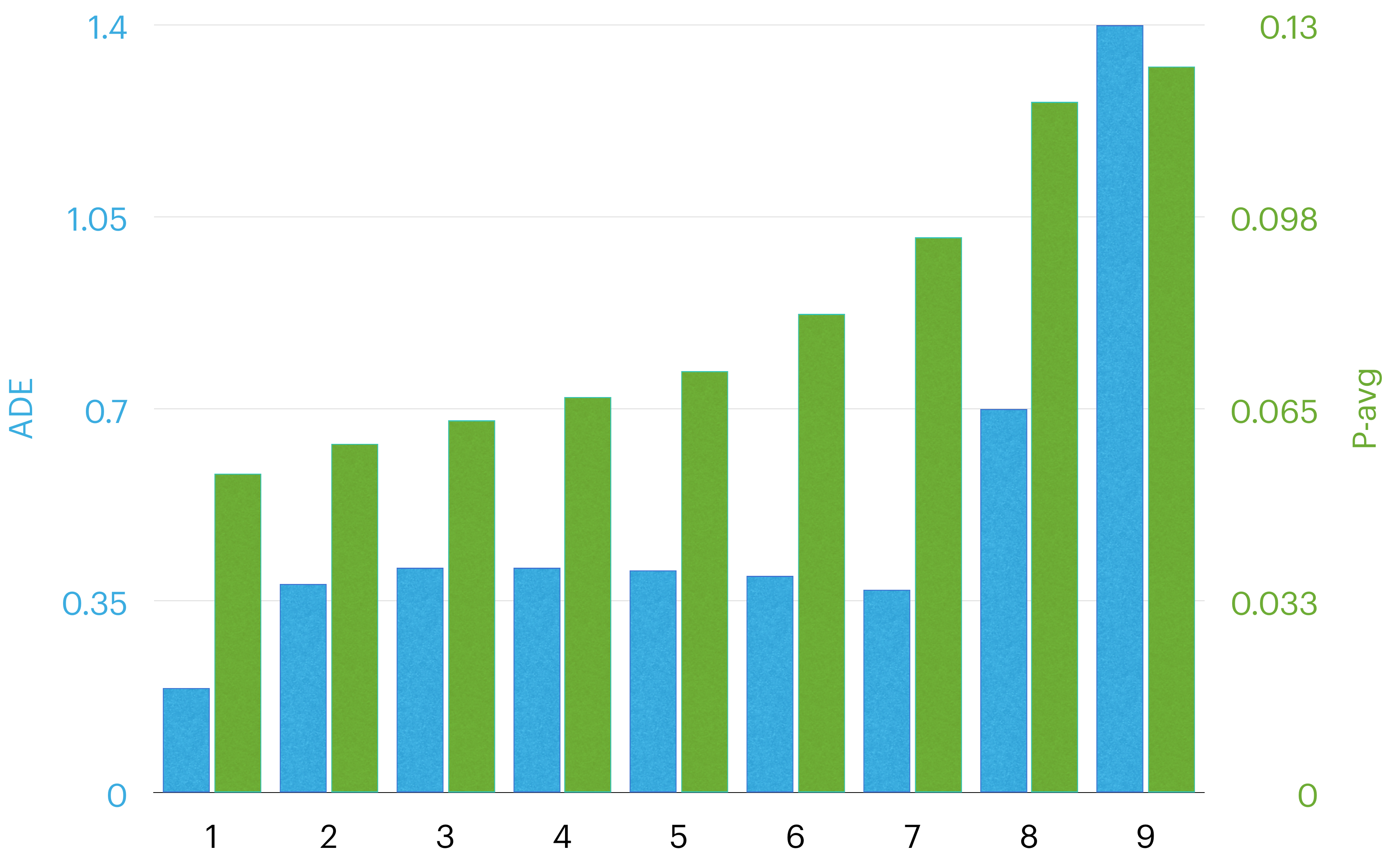}
\end{center}
   \caption{Two overlaid figures analyzing the impact of different timesteps. The blue is the average change in the predictions by adding a random noise of size $0.2$m to different observation timesteps. 
   The green curve shows the perturbation size at each timestep in perturbations found by S-ATTack.
   }
\label{fig:histogram}
\end{figure}

\item \bl{Are different prediction timesteps equally vulnerable to the collisions?  To answer this question, we consider a scenario where two pedestrians are crossing each other at different prediction timesteps and measure the collision rate. We use S-Forces,S-LSTM and D-Pool. \Cref{fig:timesensitivity}(a) visualizes the predictions of S-LSTM and S-Forces models for one scenario. While S-LSTM is unable to avoid the collision, S-Forces decreases the prediction speed in the first points to avoid the collision. Then, we study the sensitivity of models in different timesteps with regards to collision avoidance. To this end, we run scenarios like \Cref{fig:timesensitivity}(a),(b) 1000 times for each timestep where small Gaussian noise is added to the observations. The result are shown in a heatmap in \Cref{fig:timesensitivity}(c). First, a high collision rate is observed for both models revealing their weaknesses in capturing agent-agent interactions. Second, each model has a different pattern of sensitivity with respect to the timestep. Initial timesteps have higher collision rate since it is more difficult to change the manoeuvre in the first timesteps. While smaller collision rates are expected for later timesteps, in both models, the collision rate does not monotonically decrease. This shows that the models are more vulnerable in the middle timesteps. This can be due to the biases in the data or model structure.
}

\begin{figure}[!t]
\begin{center}
\begin{subfigure}[b]{0.24\linewidth}
    \includegraphics[width=\columnwidth]{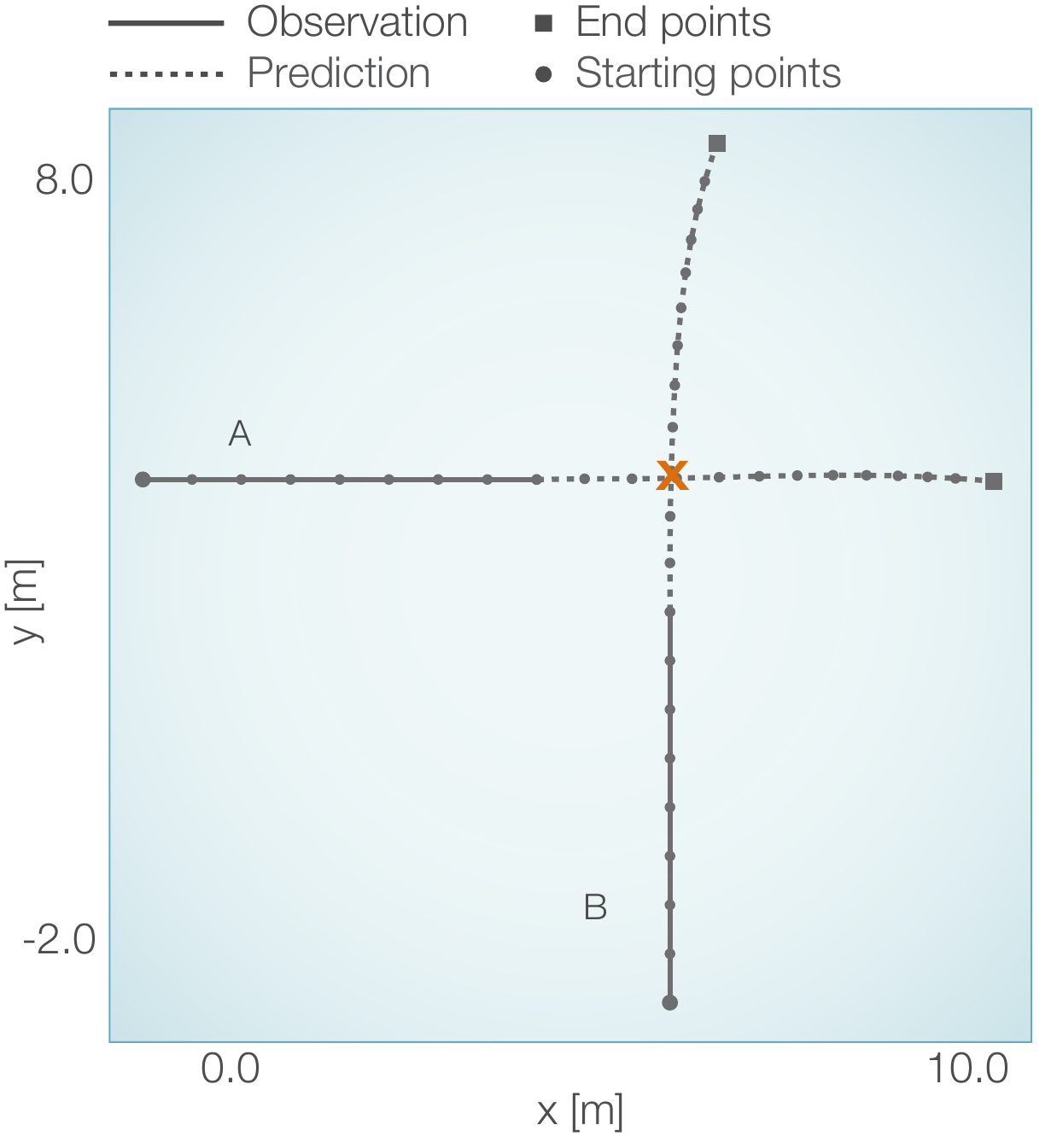}
    \caption{S-LSTM}
  \end{subfigure}
\begin{subfigure}[b]{0.24\linewidth}
    \includegraphics[width=\columnwidth]{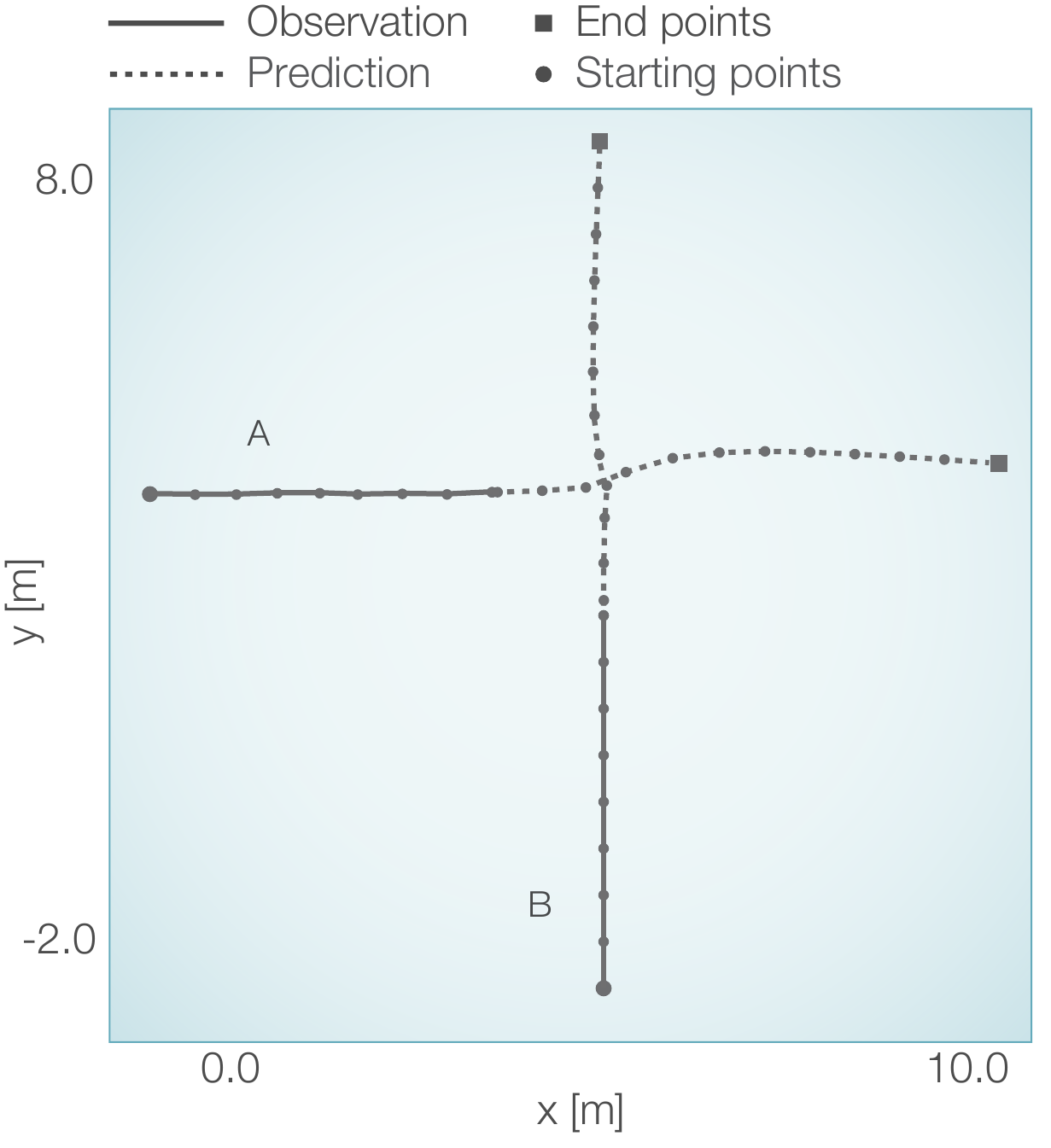}
    \caption{S-Forces}
  \end{subfigure}
\begin{subfigure}[b]{0.49\linewidth}
    \includegraphics[width=\columnwidth]{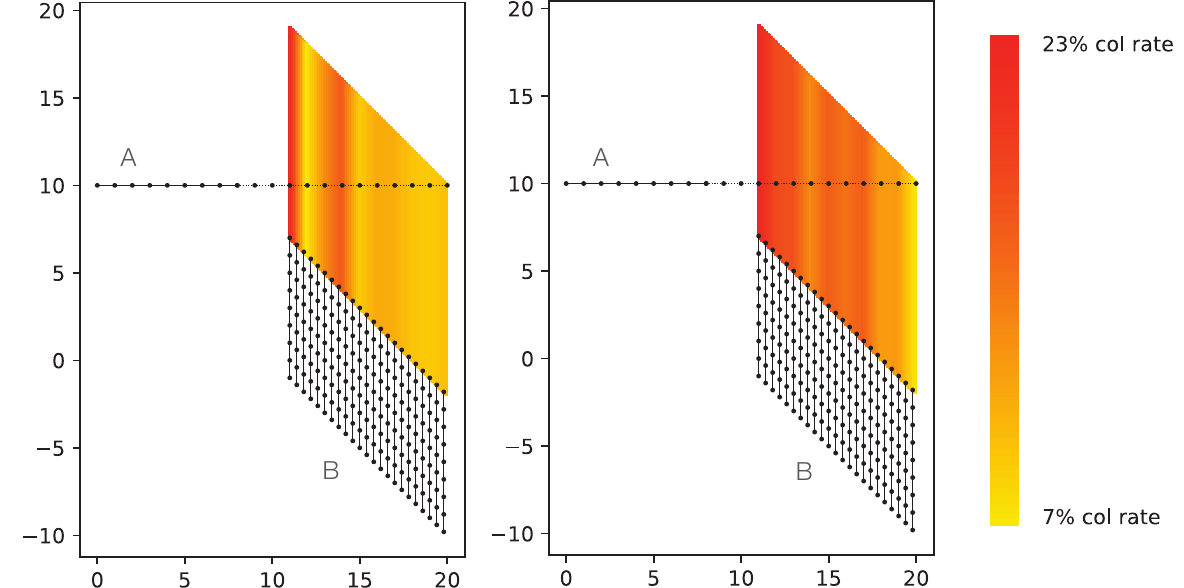}
    \caption{Prediction Heatmap}
  \end{subfigure}
\end{center}
   \caption{\bl{Collision rate analysis for different timesteps. We cross two pedestrians A,B to see if the prediction models can avoid the facing collision. (a,b) We visualize the performance of S-LSTM and S-Forces models in a scenario where two agents cross in the third prediction timestep. S-LSTM is not avoiding the collision whereas S-Forces avoids it due to the hand-crafted knowledge of collision avoidance. (c) Analysis of collision rates of two data-driven model, D-Pool (left) and S-LSTM (right) on different timesteps. We cross two pedestrians A,B on different timesteps and report the collision rate for each of them. Dark yellow is the observation sequence for both agents and green is the prediction of A. The red and yellow colors show high and low collision rates, respectively. 
   We observe different patterns for each model. We do not visualize the first two prediction timesteps as the model is unable to change the manoeuvre instantly. 
  } }
\label{fig:timesensitivity}
\end{figure}

\item \bl{Our attack perturbs only the candidate agent to achieve a collision. The same method can be employed to perturb other agents. \Cref{fig:shocking} shows three different scenarios in a scene where a collision occurs by perturbing different agents. However, collision types of \Cref{fig:shocking}(b) and \Cref{fig:shocking}(c) are not considered in our method and can be addressed as a future work.}

\item \bl{The perturbations are calculated through a joint optimization problem and there is no constraint on the smoothness of the perturbations. To study the impact of our adversarial attack in the presence of a smoothing filter, a 4th order polynomial is fit to smooth the trajectories according to \cite{becker2018evaluation}. In the first and second rows of \Cref{tab:smooth}, the performance of the original model and with S-attack is observed. Next, we add the smoothing in defense (3rd row) and in both defense and attack (4th row).
Smoothing in defense makes the final calculated perturbations smooth before feeding to the predictor. Smoothing the trajectory reduces collision but still the attack is highly effective. In the third experiment, we include the smoothing in the attack optimization algorithm where the perturbation is smoothed after each optimization iteration. Therefore, S-attack finds a smooth perturbation which shows higher collision rate compared to the previous experiment. Note that CR is lower than the original experiment which is because of the additional polynomial constraint enforced that reduces the search space.

\begin{table}[!t]
\begin{center}
\begin{tabular}{|l|c|c|}
\hline
Attacks & CR~[$\%$] $\downarrow$ \\ 
\hline\hline
D-pool  & $7.3$ \\ 
D-pool + S-attack  & $88.0$ \\ 
D-pool + S-attack + Smoothing in defense & $61.6$ \\ 
D-pool + S-attack with smoothing + Smoothing in defense & $72.0$ \\ 

\hline
\end{tabular}
\end{center}
\caption{\bl{Leveraging smoothing function in the attack. Smoothing in defense makes the perturbed observation smooth before feeding to the predictor which can be considered as a defense method against the attack. S-attack with smoothing includes the smoothing function in the optimization problem to achieve a smoothed perturbation.}}
\label{tab:smooth}
\end{table}
}  

\begin{figure}[!t]
\centering
\begin{subfigure}[b]{0.32\linewidth}
    \centering\includegraphics[width=130pt]{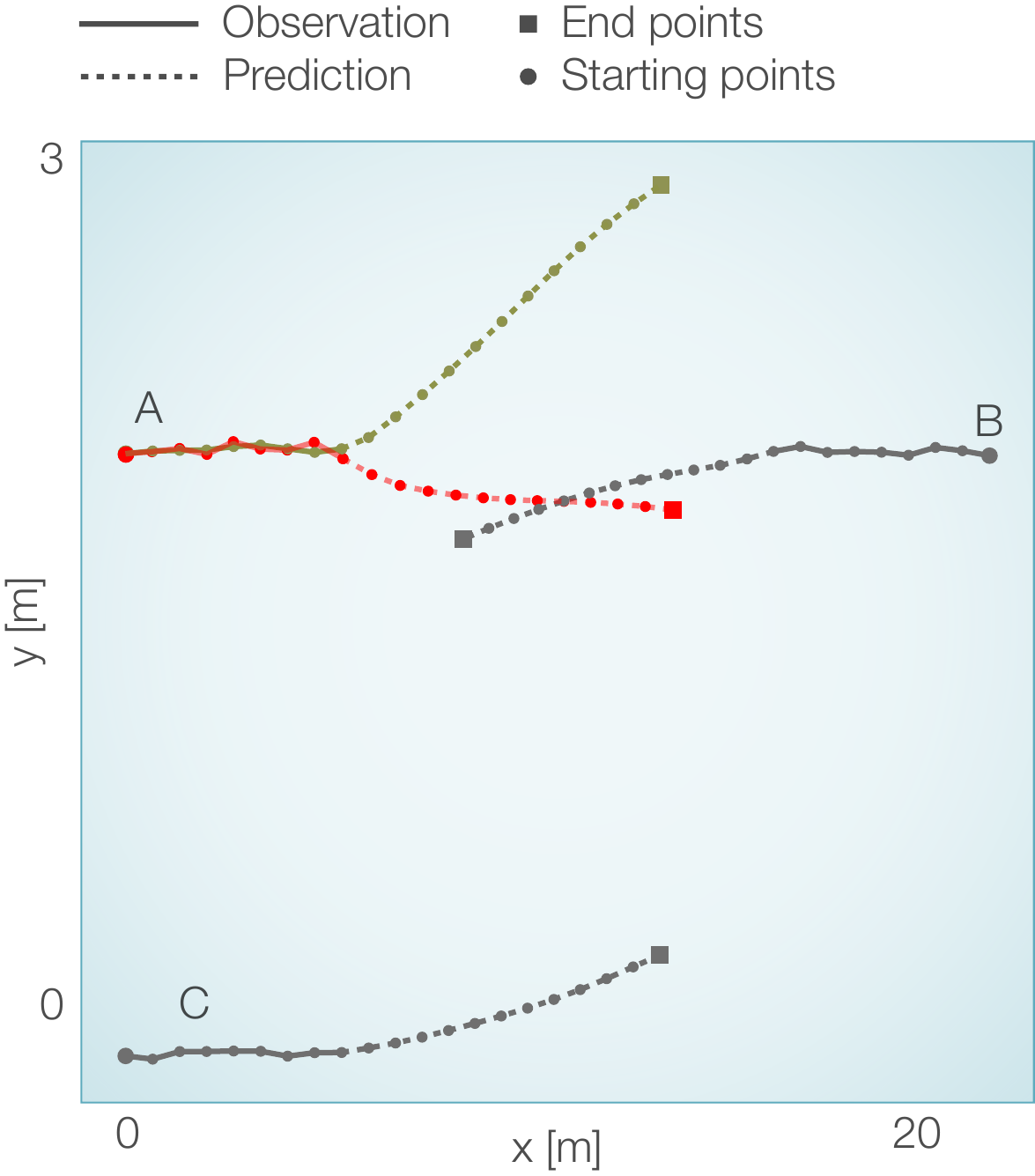}
    \caption{Agent A perturbed}
  \end{subfigure}
  \begin{subfigure}[b]{0.32\linewidth}
    \centering\includegraphics[width=130pt]{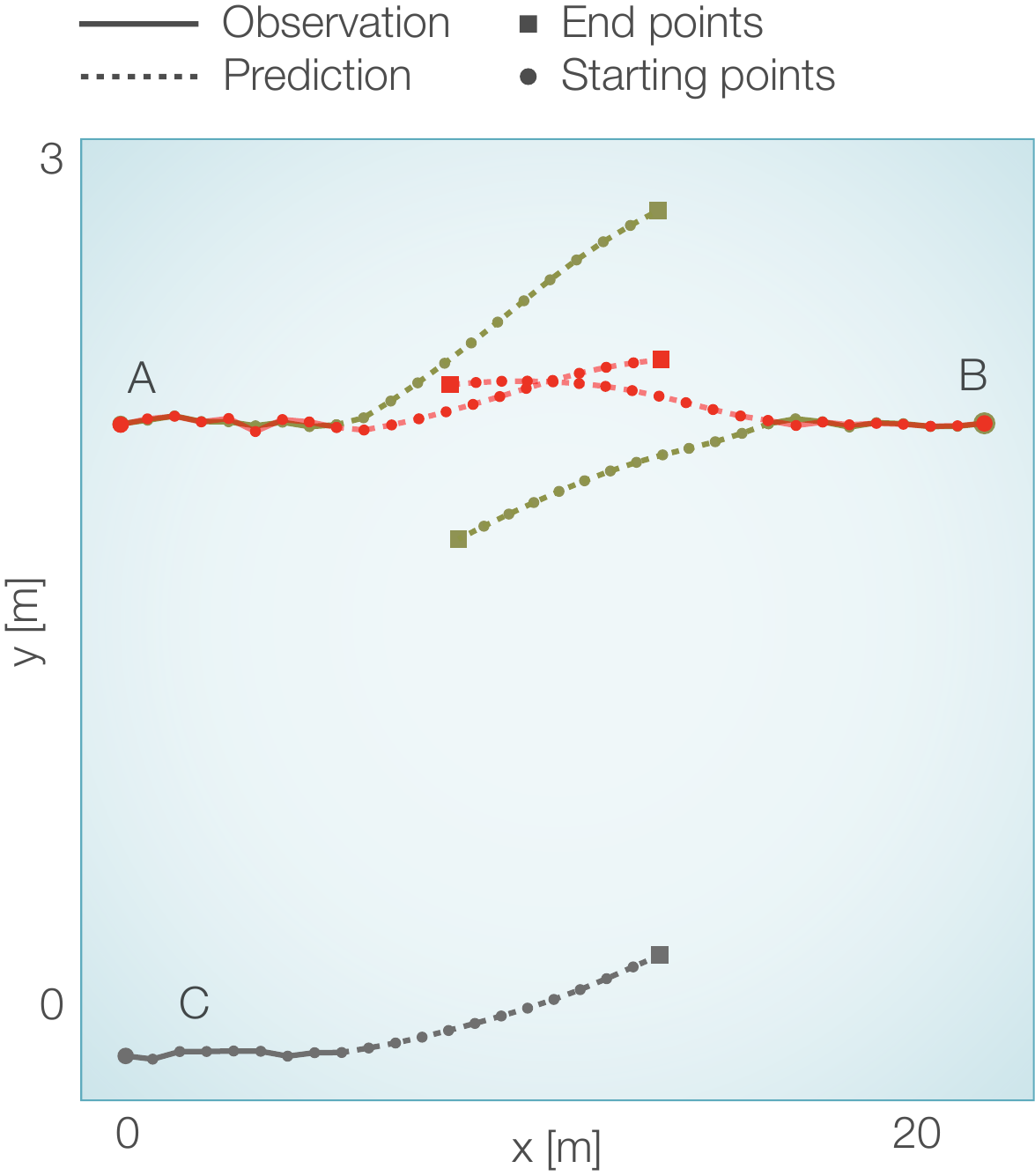}
    \caption{Agent A and B perturbed}
  \end{subfigure}
  \begin{subfigure}[b]{0.32\linewidth}
    \centering\includegraphics[width=130pt]{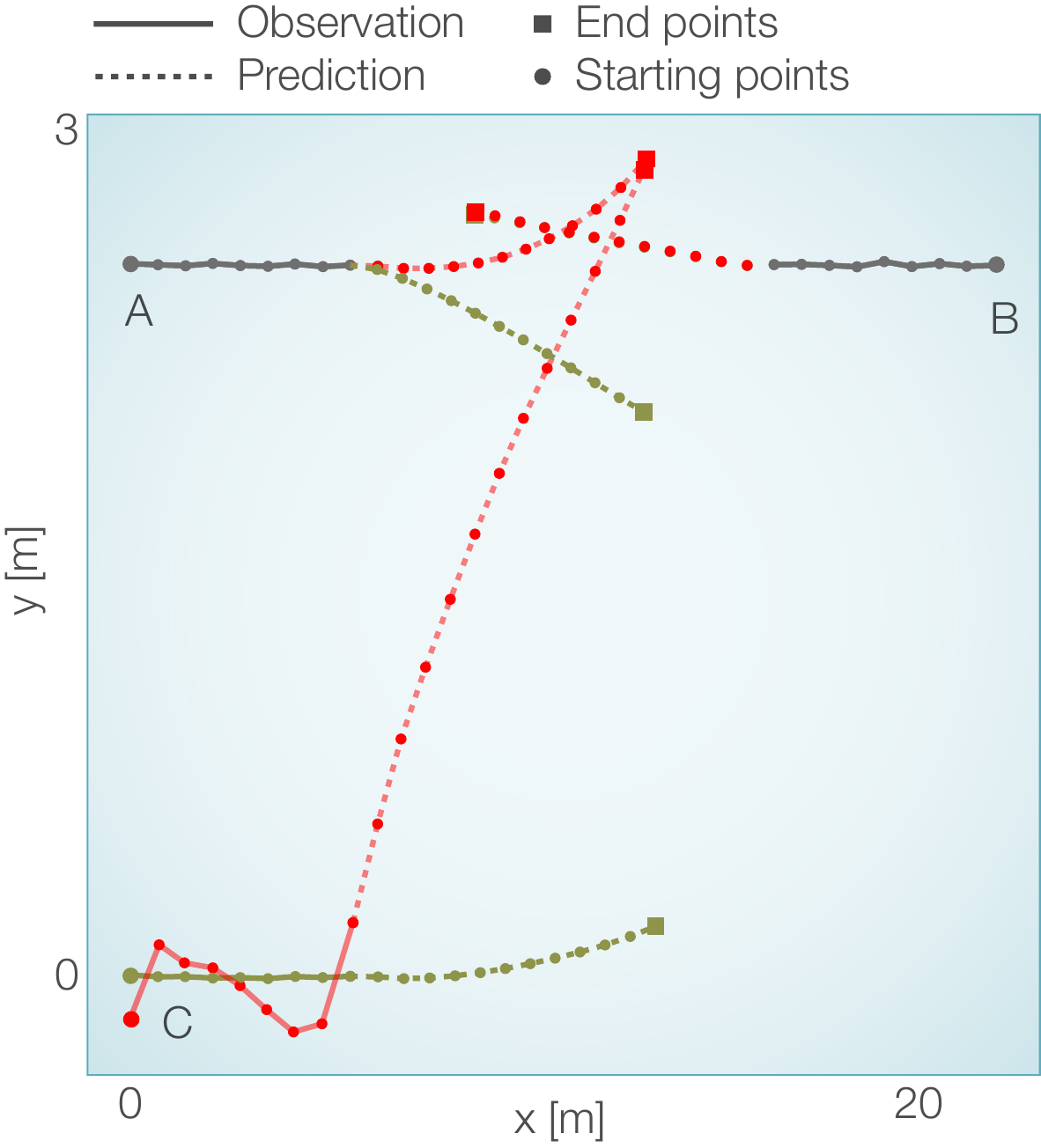}
    \caption{Agent C perturbed}
  \end{subfigure}

    \caption{\bl{Perturbing different agents in a scenario to achieve collision between agents A and B. (a) Perturbing A which is the candidate agent. (b) Perturbing both A and the other agent B that is colliding with. (c) Perturbing C as a third agent who is not colliding with A but affects the predictions of A and B. In all cases, a collision between A and B is predicted. This shows that S-attack is effective for other problem formulations.}}
\label{fig:shocking}
\end{figure}

\end{enumerate}

\section{Conclusion and \bl{future works}}
In this work, we studied the robustness properties of trajectory prediction models in terms of social understanding under adversarial attack. We introduce our Socially-ATTended ATTack (S-ATTack) to cause collisions in state-of-the-art prediction models with small perturbations. 
Adversarial training using S-ATTack can not only make the models more robust against adversarial attacks, but also reduce the collision rate and hence, improve their social understanding. This paper sheds light on the common weaknesses of trajectory prediction models opening a window towards their social understandings.

\bl{Our work is a first step that highlights the lack of social understanding in the models. 
The attack method can be extended by considering the velocity that a collision occurs with, or considering other notions of social behavior such as grouping.
To approach socially-aware predictors, we believe that the field lacks two main components. First, using  more socially-related metrics instead of ADE/FDE for training and evaluating. This is an area of research where metrics like the time-to-collision metric \cite{karamouzas2014universal}, are required to be studied more.  Second, supervised learning in the absence of proper inductive biases cannot learn social behaviors properly especially with the current small-scale datasets. One direction to address the challenge is to combine physics-based models and neural networks. Also, having more challenging and large-scale datasets can help the models learn the bias by themselves. 
}

\section{Acknowledgments} 
 This project was funded by Honda R\&D Co., Ltd and from the European Union's Horizon 2020 research and innovation programme under the Marie Sklodowska-Curie grant agreement No 754354.

\newpage

{\small
\bibliographystyle{ieee_fullname}
\bibliography{references}

\begin{thebibliography}{10}\itemsep=-1pt

\bibitem{alahi2016sociallstm}
Alexandre Alahi, Kratarth Goel, Vignesh Ramanathan, Alexandre Robicquet, Li
  Fei-Fei, and Silvio Savarese.
\newblock Social lstm: Human trajectory prediction in crowded spaces.
\newblock In {\em Proceedings of the IEEE/CVF conference on Computer Vision and
  Pattern Recognition (CVPR)}, pages 961--971, 2016.

\bibitem{alahi2017sociallstmbook}
Alexandre Alahi, Vignesh Ramanathan, Kratarth Goel, Alexandre Robicquet, Amir~A
  Sadeghian, Li Fei-Fei, and Silvio Savarese.
\newblock Learning to predict human behavior in crowded scenes.
\newblock In {\em Group and Crowd Behavior for Computer Vision}, pages
  183--207. Elsevier, 2017.

\bibitem{amirian2019social}
Javad Amirian, Jean-Bernard Hayet, and Julien Pettr{\'e}.
\newblock Social ways: Learning multi-modal distributions of pedestrian
  trajectories with gans.
\newblock In {\em IEEE/CVF conference on Computer Vision and Pattern
  Recognition (CVPR) Workshops}, pages 0--0, 2019.

\bibitem{antonini2006discrete}
Gianluca Antonini, Michel Bierlaire, and Mats Weber.
\newblock Discrete choice models of pedestrian walking behavior.
\newblock {\em Transportation Research Part B: Methodological}, 40(8):667--687,
  2006.

\bibitem{bahari2021injecting}
Mohammadhossein Bahari, Ismail Nejjar, and Alexandre Alahi.
\newblock Injecting knowledge in data-driven vehicle trajectory predictors.
\newblock {\em Transportation Research Part C: Emerging Technologies},
  128:103010, 2021.

\bibitem{ballet2019imperceptible}
Vincent Ballet, Xavier Renard, Jonathan Aigrain, Thibault Laugel, Pascal
  Frossard, and Marcin Detyniecki.
\newblock Imperceptible adversarial attacks on tabular data.
\newblock {\em arXiv preprint arXiv:1911.03274}, 2019.

\bibitem{bartoli2018context}
Federico Bartoli, Giuseppe Lisanti, Lamberto Ballan, and Alberto Del~Bimbo.
\newblock Context-aware trajectory prediction.
\newblock In {\em 2018 24th International Conference on Pattern Recognition
  (ICPR)}, pages 1941--1946. IEEE, 2018.

\bibitem{becker2018evaluation}
Stefan Becker, Ronny Hug, Wolfgang H{\"u}bner, and Michael Arens.
\newblock An evaluation of trajectory prediction approaches and notes on the
  trajnet benchmark.
\newblock {\em arXiv preprint arXiv:1805.07663}, 2018.

\bibitem{behjati2019universal}
Melika Behjati, Seyed-Mohsen Moosavi-Dezfooli, Mahdieh~Soleymani Baghshah, and
  Pascal Frossard.
\newblock Universal adversarial attacks on text classifiers.
\newblock In {\em IEEE International Conference on Acoustics, Speech and Signal
  Processing (ICASSP)}, pages 7345--7349. IEEE, 2019.

\bibitem{bennewitz2002learning}
Maren Bennewitz, Wolfram Burgard, and Sebastian Thrun.
\newblock Learning motion patterns of persons for mobile service robots.
\newblock In {\em Proceedings of the IEEE International Conference on Robotics
  and Automation}, volume~4, pages 3601--3606. IEEE, 2002.

\bibitem{bertoni2020monstereo}
Lorenzo Bertoni, Sven Kreiss, Taylor Mordan, and Alexandre Alahi.
\newblock Monstereo: When monocular and stereo meet at the tail of 3d human
  localization.
\newblock In {\em International Conference on Robotics and Automation (ICRA)},
  2021.

\bibitem{blue2001cellular}
Victor~J Blue and Jeffrey~L Adler.
\newblock Cellular automata microsimulation for modeling bi-directional
  pedestrian walkways.
\newblock {\em Transportation Research Part B: Methodological}, 35(3):293--312,
  2001.

\bibitem{bouhsain2020pedestrian}
Smail~Ait Bouhsain, Saeed Saadatnejad, and Alexandre Alahi.
\newblock Pedestrian intention prediction: A multi-task perspective.
\newblock {\em European Association for Research in Transportation (hEART)},
  2020.

\bibitem{burstedde2001simulation}
Carsten Burstedde, Kai Klauck, Andreas Schadschneider, and Johannes Zittartz.
\newblock Simulation of pedestrian dynamics using a two-dimensional cellular
  automaton.
\newblock {\em Physica A: Statistical Mechanics and its Applications},
  295(3-4):507--525, 2001.

\bibitem{chavdarova2018wildtrack}
Tatjana Chavdarova, Pierre Baqu{\'e}, St{\'e}phane Bouquet, Andrii Maksai, Cijo
  Jose, Timur~M. Bagautdinov, Louis Lettry, Pascal Fua, Luc~Van Gool, and
  François Fleuret.
\newblock Wildtrack: A multi-camera hd dataset for dense unscripted pedestrian
  detection.
\newblock {\em Prooceedings of the IEEE/CVF Conference on Computer Vision and
  Pattern Recognition (CVPR)}, pages 5030--5039, 2018.

\bibitem{chen2017decentralized}
Yu~Fan Chen, Miao Liu, Michael Everett, and Jonathan~P How.
\newblock Decentralized non-communicating multiagent collision avoidance with
  deep reinforcement learning.
\newblock In {\em 2017 IEEE international conference on robotics and automation
  (ICRA)}, pages 285--292. IEEE, 2017.

\bibitem{deng2020joint}
Wenlong Deng, Lorenzo Bertoni, Sven Kreiss, and Alexandre Alahi.
\newblock Joint human pose estimation and stereo 3d localization.
\newblock In {\em IEEE International Conference on Robotics and Automation
  (ICRA)}, pages 2324--2330. IEEE, 2020.

\bibitem{duives2013state}
Dorine~C Duives, Winnie Daamen, and Serge~P Hoogendoorn.
\newblock State-of-the-art crowd motion simulation models.
\newblock {\em Transportation research part C: emerging technologies},
  37:193--209, 2013.

\bibitem{fawaz2019adversarial}
Hassan~Ismail Fawaz, Germain Forestier, Jonathan Weber, Lhassane Idoumghar, and
  Pierre-Alain Muller.
\newblock Adversarial attacks on deep neural networks for time series
  classification.
\newblock In {\em 2019 International Joint Conference on Neural Networks
  (IJCNN)}, pages 1--8. IEEE, 2019.

\bibitem{goodfellow2014explaining}
Ian~J Goodfellow, Jonathon Shlens, and Christian Szegedy.
\newblock Explaining and harnessing adversarial examples.
\newblock {\em arXiv preprint arXiv:1412.6572}, 2014.

\bibitem{gupta2018socialgan}
Agrim Gupta, Justin Johnson, Li Fei-Fei, Silvio Savarese, and Alexandre Alahi.
\newblock Social gan: Socially acceptable trajectories with generative
  adversarial networks.
\newblock In {\em Proceedings of the IEEE/CVF conference on Computer Vision and
  Pattern Recognition}, pages 2255--2264, 2018.

\bibitem{helbing1998social-forces}
Dirk Helbing and Peter Molnar.
\newblock Social force model for pedestrian dynamics.
\newblock {\em Physical Review E}, 51, 05 1998.

\bibitem{ivanovic2019trajectron}
Boris Ivanovic and Marco Pavone.
\newblock The trajectron: Probabilistic multi-agent trajectory modeling with
  dynamic spatiotemporal graphs.
\newblock In {\em Proceedings of the IEEE International Conference on Computer
  Vision (ICCV)}, pages 2375--2384, 2019.

\bibitem{karamouzas2014universal}
Ioannis Karamouzas, Brian Skinner, and Stephen~J Guy.
\newblock Universal power law governing pedestrian interactions.
\newblock {\em Physical review letters}, 113(23):238701, 2014.

\bibitem{karim2020adversarial}
Fazle Karim, Somshubra Majumdar, and Houshang Darabi.
\newblock Adversarial attacks on time series.
\newblock {\em IEEE transactions on pattern analysis and machine intelligence},
  2020.

\bibitem{kosaraju2019socialbigat}
Vineet Kosaraju, Amir Sadeghian, Roberto Mart{\'\i}n-Mart{\'\i}n, Ian Reid,
  Hamid Rezatofighi, and Silvio Savarese.
\newblock Social-bigat: Multimodal trajectory forecasting using bicycle-gan and
  graph attention networks.
\newblock In {\em Advances in Neural Information Processing Systems (NeurIPS)},
  pages 137--146, 2019.

\bibitem{kothari2020human}
Parth Kothari, Sven Kreiss, and Alexandre Alahi.
\newblock Human trajectory forecasting in crowds: A deep learning perspective.
\newblock {\em IEEE Transactions on Intelligent Transportation Systems}, 2021.

\bibitem{lerner2007ucy}
Alon Lerner, Yiorgos Chrysanthou, and Dani Lischinski.
\newblock Crowds by example.
\newblock {\em Comput. Graph. Forum}, 26:655--664, 2007.

\bibitem{li2020social}
Jiachen Li, Hengbo Ma, Zhihao Zhang, and Masayoshi Tomizuka.
\newblock Social-wagdat: Interaction-aware trajectory prediction via
  wasserstein graph double-attention network.
\newblock {\em arXiv preprint arXiv:2002.06241}, 2020.

\bibitem{liu2020adversarial}
Jian Liu, Naveed Akhtar, and Ajmal Mian.
\newblock Adversarial attack on skeleton-based human action recognition.
\newblock {\em IEEE Transactions on Neural Networks and Learning Systems},
  2020.

\bibitem{madry2017pgd}
Aleksander Madry, Aleksandar Makelov, Ludwig Schmidt, Dimitris Tsipras, and
  Adrian Vladu.
\newblock Towards deep learning models resistant to adversarial attacks.
\newblock {\em arXiv preprint arXiv:1706.06083}, 2017.

\bibitem{madry2017towards}
Aleksander Madry, Aleksandar Makelov, Ludwig Schmidt, Dimitris Tsipras, and
  Adrian Vladu.
\newblock Towards deep learning models resistant to adversarial attacks.
\newblock {\em arXiv preprint arXiv:1706.06083}, 2017.

\bibitem{mangalam2020pecnet}
Karttikeya Mangalam, Harshayu Girase, Shreyas Agarwal, Kuan-Hui Lee, Ehsan
  Adeli, Jitendra Malik, and Adrien Gaidon.
\newblock It is not the journey but the destination: Endpoint conditioned
  trajectory prediction.
\newblock In {\em European Conference on Computer Vision (ECCV)}, pages
  759--776. Springer, 2020.

\bibitem{mohamed2020stgcnn}
Abduallah Mohamed, Kun Qian, Mohamed Elhoseiny, and Christian Claudel.
\newblock Social-stgcnn: A social spatio-temporal graph convolutional neural
  network for human trajectory prediction.
\newblock In {\em Proceedings of the IEEE/CVF Conference on Computer Vision and
  Pattern Recognition (CVPR)}, pages 14424--14432, 2020.

\bibitem{moosavi2016deepfool}
Seyed-Mohsen Moosavi-Dezfooli, Alhussein Fawzi, and Pascal Frossard.
\newblock Deepfool: a simple and accurate method to fool deep neural networks.
\newblock In {\em Proceedings of the IEEE/CVF Conference on Computer Vision and
  Pattern Recognition (CVPR)}, pages 2574--2582, 2016.

\bibitem{nikhil2018convolutional}
Nishant Nikhil and Brendan Tran~Morris.
\newblock Convolutional neural network for trajectory prediction.
\newblock In {\em European Conference on Computer Vision (ECCV) Workshops},
  2018.

\bibitem{NEURIPS2020_1ea97de8}
Guillermo Ortiz-Jimenez, Apostolos Modas, Seyed-Mohsen Moosavi, and Pascal
  Frossard.
\newblock Hold me tight! influence of discriminative features on deep network
  boundaries.
\newblock In H. Larochelle, M. Ranzato, R. Hadsell, M.~F. Balcan, and H. Lin,
  editors, {\em Advances in Neural Information Processing Systems (NeurIPS)},
  volume~33, pages 2935--2946. Curran Associates, Inc., 2020.

\bibitem{pellegrini2010eth}
Stefano Pellegrini, Andreas Ess, and Luc~Van Gool.
\newblock Improving data association by joint modeling of pedestrian
  trajectories and groupings.
\newblock In {\em European Conference on Computer Vision (ECCV)}, pages
  452--465. Springer, 2010.

\bibitem{robicquet2016learning}
Alexandre Robicquet, Amir Sadeghian, Alexandre Alahi, and Silvio Savarese.
\newblock Learning social etiquette: Human trajectory understanding in crowded
  scenes.
\newblock In {\em European Conference on Computer Vision (ECCV)}, pages
  549--565. Springer, 2016.

\bibitem{robin2009specification}
Th Robin, Gianluca Antonini, Michel Bierlaire, and Javier Cruz.
\newblock Specification, estimation and validation of a pedestrian walking
  behavior model.
\newblock {\em Transportation Research Part B: Methodological}, 43(1):36--56,
  2009.

\bibitem{sadeghian2019sophie}
Amir Sadeghian, Vineet Kosaraju, Ali Sadeghian, Noriaki Hirose, Hamid
  Rezatofighi, and Silvio Savarese.
\newblock Sophie: An attentive gan for predicting paths compliant to social and
  physical constraints.
\newblock In {\em Proceedings of the IEEE/CVF conference on Computer Vision and
  Pattern Recognition (CVPR)}, pages 1349--1358, 2019.

\bibitem{SEER2014212}
Stefan Seer, Norbert Brändle, and Carlo Ratti.
\newblock Kinects and human kinetics: A new approach for studying pedestrian
  behavior.
\newblock {\em Transportation Research Part C: Emerging Technologies},
  48:212--228, 2014.

\bibitem{su2019onepixelattack}
Jiawei Su, Danilo~Vasconcellos Vargas, and Kouichi Sakurai.
\newblock One pixel attack for fooling deep neural networks.
\newblock {\em IEEE Transactions on Evolutionary Computation}, 23(5):828--841,
  2019.

\bibitem{szegedy2013intriguing}
Christian Szegedy, Wojciech Zaremba, Ilya Sutskever, Joan Bruna, Dumitru Erhan,
  Ian Goodfellow, and Rob Fergus.
\newblock Intriguing properties of neural networks.
\newblock {\em arXiv preprint arXiv:1312.6199}, 2013.

\bibitem{tsipras2018robustness}
Dimitris Tsipras, Shibani Santurkar, Logan Engstrom, Alexander Turner, and
  Aleksander Madry.
\newblock Robustness may be at odds with accuracy.
\newblock {\em arXiv preprint arXiv: 1805.12152}, 2018.

\bibitem{HeidenBMVC2019}
T. van~der Heiden, N.~S. Nagaraja, C. Weiss, and E. Gavves.
\newblock Safecritic: Collision-aware trajectory prediction.
\newblock In {\em British Machine Vision Conference Workshop}, 2019.

\bibitem{vemula2018socialattention}
Anirudh Vemula, Katharina Muelling, and Jean Oh.
\newblock Social attention: Modeling attention in human crowds.
\newblock In {\em 2018 IEEE international Conference on Robotics and Automation
  (ICRA)}, pages 1--7. IEEE, 2018.

\bibitem{vizzari2015agent}
Giuseppe Vizzari, Lorenza Manenti, Kazumichi Ohtsuka, and Kenichiro Shimura.
\newblock An agent-based pedestrian and group dynamics model applied to
  experimental and real-world scenarios.
\newblock {\em Journal of Intelligent Transportation Systems}, 19(1):32--45,
  2015.

\bibitem{yu2020spatio}
Cunjun Yu, Xiao Ma, Jiawei Ren, Haiyu Zhao, and Shuai Yi.
\newblock Spatio-temporal graph transformer networks for pedestrian trajectory
  prediction.
\newblock In {\em European Conference on Computer Vision (ECCV)}, pages
  507--523. Springer, 2020.

\bibitem{zeng2019end}
Wenyuan Zeng, Wenjie Luo, Simon Suo, Abbas Sadat, Bin Yang, Sergio Casas, and
  Raquel Urtasun.
\newblock End-to-end interpretable neural motion planner.
\newblock In {\em Proceedings of the IEEE/CVF Conference on Computer Vision and
  Pattern Recognition (CVPR)}, pages 8660--8669, 2019.

\bibitem{zhang2020adversarial}
Wei~Emma Zhang, Quan~Z Sheng, Ahoud Alhazmi, and Chenliang Li.
\newblock Adversarial attacks on deep-learning models in natural language
  processing: A survey.
\newblock {\em ACM Transactions on Intelligent Systems and Technology (TIST)},
  11(3):1--41, 2020.

\end{thebibliography}
}

\end{document}